\documentclass[journal]{IEEEtran}
\IEEEoverridecommandlockouts 

\usepackage[numbers]{natbib}
\usepackage{graphics}
\usepackage{caption}
\usepackage{epsfig}
\usepackage{times}
\usepackage{url}
\usepackage{algorithmic}
\usepackage{amsmath}
\usepackage{array}
\usepackage[mathscr]{euscript}
\usepackage{amssymb}
\usepackage{nomencl}
\usepackage{textcomp}
\usepackage{tabularx}
\usepackage{multirow}
\usepackage{lipsum}
\usepackage{xcolor}
\usepackage{gensymb}
\usepackage{enumitem}
\newcommand{\RN}[1]{%
  \textup{\uppercase\expandafter{\romannumeral#1}}%
}

\usepackage{xspace}
\DeclareRobustCommand\onedot{\futurelet\@let@token\@onedot}
\def\@onedot{\ifx\@let@token.\else.\null\fi\xspace}

\usepackage[ruled,linesnumbered]{algorithm2e}
\usepackage{float}

\title{\LARGE \bf
Human-Robot Interaction via a Joint-Initiative Supervised Autonomy (JISA) Framework.}

\author{Abbas Sidaoui, Naseem Daher, and Daniel Asmar
\thanks{Abbas Sidaoui, Naseem Daher, and Daniel Asmar are with the Vision and Robotics Lab, American University of Beirut, Beirut, P.O. Box 11-0236 Lebanon. E-mails: (\{ams108, nd38, da20\}@aub.edu.lb)}
}

\usepackage{xspace}
\makeatletter
\DeclareRobustCommand\onedot{\futurelet\@let@token\@onedot}
\def\@onedot{\ifx\@let@token.\else.\null\fi\xspace}

\makeatother

\begin{document}
\maketitle
\thispagestyle{empty}
\pagestyle{empty}

\begin{abstract}
In this paper, we propose and validate a Joint-Initiative Supervised Autonomy (JISA) framework for Human-Robot Interaction (HRI), in which a robot maintains a measure of its self-confidence (SC) while performing a task, and only prompts the human supervisor for help when its SC drops. At the same time, during task execution, a human supervisor can intervene in the task being performed, based on his/her Situation Awareness (SA).  To evaluate the applicability and utility of JISA, it is implemented on two different HRI tasks: grid-based collaborative simultaneous localization and mapping (SLAM) and automated jigsaw puzzle reconstruction. Augmented Reality (AR) (for SLAM) and two-dimensional graphical user interfaces (GUI) (for puzzle reconstruction) are custom-designed to enhance human SA and allow intuitive interaction between the human and the agent. The superiority of the JISA framework is demonstrated in experiments. In SLAM, the superior maps produced by JISA preclude the need for post processing of any SLAM stock maps; furthermore, JISA reduces the required mapping time by approximately 50 percent versus traditional approaches. In automated puzzle reconstruction, the JISA framework outperforms both fully autonomous solutions, as well as those resulting from on-demand human intervention prompted by the agent.
\end{abstract}

\section{Introduction}

When automation was first introduced, its use was limited to well-structured and controlled environments in which each automated agent performed a very specific task \cite{c1}. Nowadays, with the rapid advancements in automation and the decreasing cost of hardware, robots are being used to automate more tasks in unstructured and dynamic environments. However, as the tasks become more generic and the environments become more unstructured, full autonomy is challenged by the limited perception, cognition, and execution capabilities of robots \cite{mot_30}. Furthermore, when robots operate in full autonomy, the noise in the real world increases the possibility of making mistakes \cite{mot_32}, which could be due to certain decisions or situations that the robot is uncertain about, or due to limitations in the software/hardware abilities. In some cases, the robot is able to flag that there might be a problem (\textit{e.g.}, high uncertainty in object detection or inability to reach a manipulation goal); while in other cases, the robot may not even be aware that errors exist (\textit{e.g.}, LiDAR not detecting a transparent obstacle).

In recent years, human-robot interaction has been gaining more attention from robotics researchers \cite{IntellROb4}. First, robots are being deployed to automate more tasks in close proximity with humans \cite{IntellROb1, IntellROb2}; second, the challenges facing autonomous robots may not perplex humans, owing to their superior cognition, reasoning, and problem-solving skills. This has lead researchers in the HRI community to include humans in the decision-making loop to assist and provide help, as long as autonomous systems are not fully capable of handling all types of contingencies \cite{mot_39, mot_21,  IntellROb3}. Including a human in the loop can mitigate certain challenges and limitations facing autonomy, by complementing the strength, endurance, productivity, and precision that a robot offers on one hand, with the cognition, flexibility, and problem-solving abilities of humans on the other. In fact, studies have shown that combining human and robots skills can increase the capabilities, reliability, and robustness of robots \cite{ras_1}. Thus, creating successful approaches to human-robot interaction and human-autonomy teaming is considered crucial for developing effective autonomous systems \cite{mot_42, ras_6}, and deploying robots outside of controlled and structured environments \cite{ras_7, ras_8}.

One approach to human-autonomy teaming is having a human who helps the autonomous agents when necessary. In fact, seeking and providing help to overcome challenges in performing tasks has been studied by human-psychology researchers for many decades. When uncertain about a situation, people tend to gather missing information, assess different alternatives, and avoid mistakes by seeking help from more knowledgeable persons \cite{mot_a11}; for example, sometimes junior developers consult senior colleagues to help in detecting hidden bugs in a malfunctioning code, or students ask teachers to check an assignment answer that they are not certain about. 

Help is also considered indispensable to solve problems and achieve goals that are beyond a person’s capabilities \cite{mot_a11}. When aware of their own limitations, people seek help to overcome their disadvantaged situations; for instance, a shorter employee may ask a taller colleague to hand them an object that is placed at a high shelf. In some cases, people might not be aware of the situation nor their limitations; here, another person might have to intervene and provide help to avoid problems or unwanted outcomes. For example, a supervisor at work may step in to correct a mistake that an intern is not aware of, or when a passer-by helps a blind person avoid an obstacle on the sidewalk. Moreover, exchanging help has been shown to enhance the performance and effectiveness of human teams \cite{mot_a5, mot_a8, mot_a7}, especially in teams with members having different backgrounds, experiences, and skill sets. 
\begin{figure*}[!t]
\includegraphics[width=\textwidth]{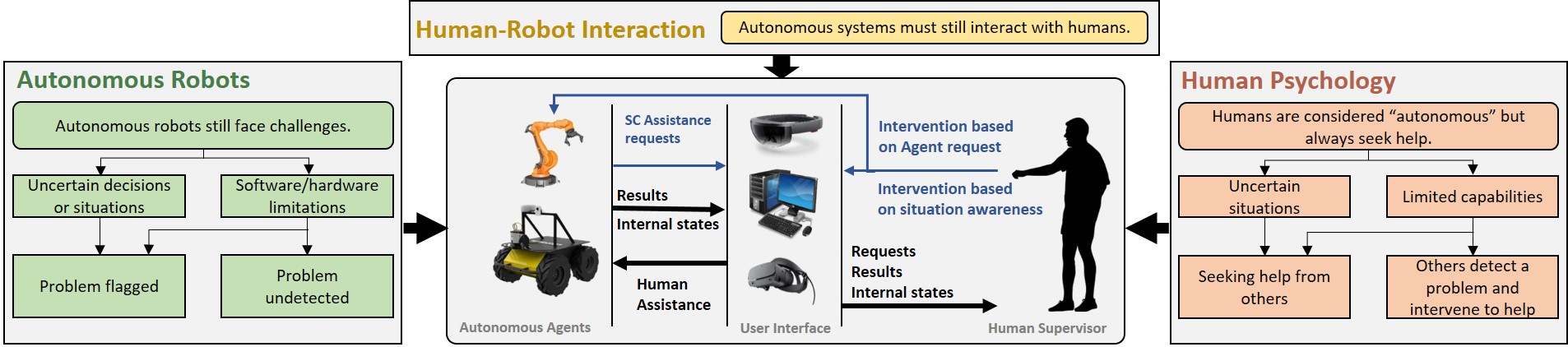}
\centering
\caption{The proposed JISA framework is motivated by needs from three areas: autonomous robots, human psychology, and human-robot interaction. The grey block in the middle shows an illustration of the proposed JISA system. The autonomous agent(s) perform the tasks autonomously and send the human supervisor their internal states, tasks results, and SC-based assistance requests through a user interface. The supervisor can assist the robot or intervene either directly or through the user interface.}
\label{fig:motivation}
\end{figure*}

With the above in mind, Fig. \ref{fig:motivation} summarizes the motivation behind our proposed JISA framework from various perspectives. First, the human psychology literature indicates that although humans think of themselves as being autonomous \cite{mot_a10}, they always seek and receive help to achieve their goals and overcome uncertain situations and limited capabilities. Second, robotics literature states that autonomous robots are still facing challenges due to limitations in hardware, software, and cognitive abilities. These challenges lead to mistakes and problems; some of which can be detected and flagged by the robot, while others may go undetected by the robot. Third, research in HRI and human-autonomy interaction illustrates that having a human to interact with and assist autonomous agents is a must. 

From the three different lines of research, we conclude that there is a need for a new HRI framework that takes advantage of the complementary skills of humans and robots on one hand, and adopt the general idea of help-seeking and helping behaviors from human-human interaction on the other hand. While state-of-the-art frameworks focus mainly on how to allocate and exchange tasks between humans and robots, there is no generalized framework, up to our knowledge, that considers including a human in the loop to support autonomous systems where a human cannot completely takeover the tasks execution. For this reason, we are proposing JISA as a framework for HRI that allows a human to supervise autonomous tasks and provide help to avoid autonomy mistakes, without a complete takeover of these tasks. Human assistance can be requested by the robot in tasks where it can reason about its SC. In tasks where the robot is not aware of its limitations, or does not have the means to measure its SC, human assistance is based on the human's SA. Moreover, the type, timing, and extent of interaction between the human supervisor and the robot are governed within JISA through a JISA-based user interface.

The main contributions of this paper are:
\begin{enumerate}
    \item Proposing a \textbf{novel} Joint-Initiative Supervised Autonomy (JISA) framework for human-robot interaction, which leverages the robot’s SC and the human's SA in order to improve the performance of automated tasks and mitigate the limitations of fully autonomous systems without having a human to completely takeover any task execution. 
    \item Demonstrating the utility and applicability of JISA by:
        \subitem a) Re-framing the works in \cite{aug_mapping, aug_SLAM, coll_SLAM} in the context of the proposed JISA framework, with the aim of showing how existing works can be \textbf{generalized} under the JISA framework.
        \subitem b) Applying the JISA framework in a \textbf{new application} that involves automated jigsaw puzzle solving. 
    
    These two applications serve as proof-of-concept (POC) to the proposed JISA framework.
    \item Experimental validation and providing preliminary empirical evidence that applying JISA outperforms full autonomy in both POC applications. We note that the experiments in section \ref{sec:JISA_puzzle} were conducted uniquely for this manuscript, while experiments in section \ref{sec:JISA_SLAM} were introduced in \cite{aug_SLAM, coll_SLAM}. No additional experiments were conducted in section \ref{sec:JISA_SLAM}, since our goal is to show how a previous approach could be generalized and re-formulated in a more systematic manner under the JISA framework.
    
\end{enumerate}

\section{Related Work}
\label{sec:related_work}
Researchers in the HRI community have been advocating bringing back humans to the automation loop through different approaches and methods. This section presents some of the most common approaches.

One of the commonly used methods for HRI is Mixed Initiative Human-Robot Interaction (MI-HRI). The term `Mixed-initiative' (MI) was first introduced in 1970 in the context of computer-assisted instruction systems \cite{c13}. This definition was followed by several others in the domain of Human-Computer Interaction (HCI) \cite{c15,c16,c17}, where MI was proposed as a method to enable both agents and humans to contribute in planning and problem-solving according to their knowledge and skills. Motivated by the previous definitions, Jiang and Arkin \cite{c14} presented the most common definition for MI-HRI, which states that `an initiative (task) is mixed only when each member of a human-robot team is authorized to intervene and seize control of it' \cite{c14}. According to this definition, the term `initiative' in MI-HRI systems refers to a task of the mission, ranging from low-level control to high-level decisions, that could be performed and seized by either the robot or the human \cite{c14}. This means that both agents (robot and human) may have tasks to perform within a mission, and both can completely takeover tasks from each other. Unlike MI, in JISA the term `initiative' is used to refer to `initiating the human assistance,' and it is considered joint since this assistance is initiated either by the autonomous agent's request or by the human supervisor. Another difference between JISA and MI-HRI is that all tasks within a JISA system are performed by the robot; albeit, human assistance is requested/provided to overcome certain challenges and limitations.

Another method that enables the contribution of both humans and agents in an automated task is `adjustable autonomy', which allows one to change the levels of autonomy within an autonomous system by the system itself, the operator, or an external agent\cite{c18}. 
This adjustment in autonomy enables a user to prevent issues by either partial assistance or complete takeover, depending on the need.  However, when the autonomy is adjusted, tasks are re-allocated between the human and the agents, and certain conditions have to be met in order to re-adjust the system back to the full-autonomy state. In our proposed framework, the tasks are not re-allocated between the human and the agent but rather the human is acting as a supervisor to assist the autonomous agent and intervene when needed. Since no additional tasks are re-allocated to the human in JISA, he/she can supervise multiple agents at once with, theoretically, less mental and physical strain as compared to re-assigning tasks to the human himself.

It is important to mention that we are not proposing JISA to replace these previous approaches, but rather to expand the spectrum of HRI to autonomous systems where it is not feasible for a human to `completely takeover' the mission tasks. Examples of such tasks are real-time localization of a mobile robot, motion planning of multiple degrees-of-freedom robotic manipulators, path planning of racing autonomous ground/aerial vehicles, among others. In fact, every HRI approach has use-cases where it performs better in. Adjustable autonomy and MI-HRI focus mainly on how to allocate functions between the robot and the human to enhance the performance of a collaborative mission. However, JISA is proposed to enhance the performance, decrease uncertainty, and extend capabilities of robots that are initially fully autonomous by including a human in the loop to assist the robot without taking over any task execution. 

Setting the Level of Robot Autonomy (LORA) was also considered by researchers as a method for HRI.	Beer \textit{et al.} \cite{c3} proposed a taxonomy for LORA that considers the perspective of HRI and the roles of both the human and the robot. This taxonomy consists of 10 levels of autonomy that are defined based on the allocation of functions between robot and human in each of the sense, plan, and act primitives. Among the different levels of autonomy included in the LORA taxonomy, there are two levels of relevance to our work: (1) shared control with robot initiative where human input is requested by the robot, and (2) shared control with human initiative where human input is based on his own judgment. Although these two levels allow for human assistance, it is limited to the sense and plan primitives. In addition, the continuum does not contain a level which allows interaction based on both human judgment and requests from the robot. To mitigate the limitations of these two LORAs, our JISA framework proposes a new level labelled as `Joint-Initiative Supervised Autonomy (JISA) Level,' stated as follows:

\textit{The robot performs all three primitives (sense, plan, act) autonomously, but can ask for human assistance based on a pre-defined internal state. In addition, the human can intervene to influence the output of any primitive when s/he finds it necessary.} 

Moreover, and unlike LORA, JISA provides a clear framework for implementation. 

 In addition to the general approaches discussed above, several application-oriented methods were presented in which human help is exploited to overcome autonomy challenges and limitations. Certain researchers focused on the idea of having the human decide when to assist the robot or change its autonomy level. For instance, systems that allow the user to switch autonomy level in a navigation task were proposed in \cite{c25, c27}. Although these methods allow the human to switch between autonomy levels, the definition of the levels and the tasks allocated to the human in each level are different and mission-specific. In \cite{ras_2}, a human operator helps the robot in enhancing its semantic knowledge by tagging objects and rooms via voice commands and a laser pointer. In such human-initiated assistance approaches, helping the robot and/or changing the autonomy level is solely based on the human judgment; thus, the human has to closely monitor the performance in order to avoid failures or decaying performance, which results in additional mental/physical workload. 

Other researchers investigated approaches where interaction and human help is based on the robot's request only. In \cite{ras_1}, a framework that allows the robot to request human help based on the value of information (VOI) theory was proposed. In this framework, the human is requested to help the robot by providing perceptual input to overcome autonomy challenges. In \cite{c22}, time and cost measures, instead of VOI, were used to decide on the need for interaction with the human. In \cite{IntellROb5} and \cite{ras_3}, human assistance is requested whenever the robot faces difficulties or unplanned situations while performing navigation tasks. Instead of requesting human help in decision-making, \cite{c23} proposed a mechanism that allows the robot to switch between different autonomy levels based on the `robot’s self-confidence’ and the modeled `human trust in the robot.’ In this approach, the robot decreases its autonomy level, and thus increases its dependency on the human operator, whenever its confidence in accurately performing the task drops. In these robot-initiated assistance approaches, the human interaction is requested either directly through an interface or by the agent changing its autonomy level; thus, any failure to detect the need for human assistance would lead to performance deterioration and potentially catastrophic consequences.

Other methods considered humans assisting robots based on both human judgment and robots requests. In \cite{ras_4}, an interactive human-robot semantic sensing framework was proposed. Through this framework, the robot can pose questions to the operator, who acts as a human sensor,  in order to increase its semantic knowledge or assure its observations in a target search mission. Moreover, the human operator can intervene at any moment to provide useful information to the robot. Here, human assistance is limited to the sensing primitive. In the context of adjustable autonomy, \cite{c26} proposed a remotely operated robot that changes its navigation autonomy level when a degradation in performance is detected. In addition, operators can choose to change autonomy levels based on their own judgment. In \cite{c_mi4}, a human can influence the robot autonomy via direct control or task assignment, albeit the proposed controller allows the robot to take actions in case of human error. Some researchers refer to MI as the ability of the robot to assist in or takeover the tasks that are initially performed by the human \cite{c_mi6}, or combine human inputs with automation recommendations to enhance performance \cite{c_mi8}. Finally, in \cite{c_mi7} the concept of MI is used to decide what task to be performed by which agent in a human-robot team. In contrast to such systems where the human is a teammate who performs tasks, the JISA framework proposes the human as a supervisor who intervenes to assist the autonomous agent as needed: either on-demand or based on SA. 

\section{Proposed Methodology}
\label{sec:meth}
In our proposed JISA framework, we consider HRI systems consisting of a human supervisor and autonomous agent(s). We define JISA as: "A Joint-Initiative Supervised Autonomy framework that allows a robot to sense, plan, and act autonomously, but can ask for human assistance based on its internal self-confidence. In addition, the human supervisor can intervene and influence the robot sensing, planning, and acting at any given moment based on his/her SA." According to this definition, the robot operates autonomously but the human supervisor is allowed to help in reducing the robot's uncertainty and increasing its capabilities through SC-based and SA-based assistance. The application of JISA in HRI systems relies on the following assumptions:

\begin{enumerate}
	\item The human is aware of the application being performed and its automation limitations. 
	\item The autonomous agent has a defined self-confidence based mechanism that it can use to prompt the user for help. Through this mechanism, the robot requests the human supervisor to approve, decline, or edit the decisions it is not certain about. Moreover, this mechanism is used to inform the user if the robot is unable to perform a task.
	\item When the autonomous agent seeks help, the human is both ready to assist and capable of providing the needed assistance.
    \item The JISA system is equipped with an intuitive communication channel (JISA-based interface) that allows the agent to ask for help and the human to assist and intervene in the tasks.
    \item The human has adequate expertise and SA, supported by the JISA-based interface, to intervene in the supervised task.
\end{enumerate}

\noindent Fig. \ref{fig:Meth_flowchart} presents the guideline for implementing JISA, including the following steps:
\begin{figure}[t]
\includegraphics[width=3 in]{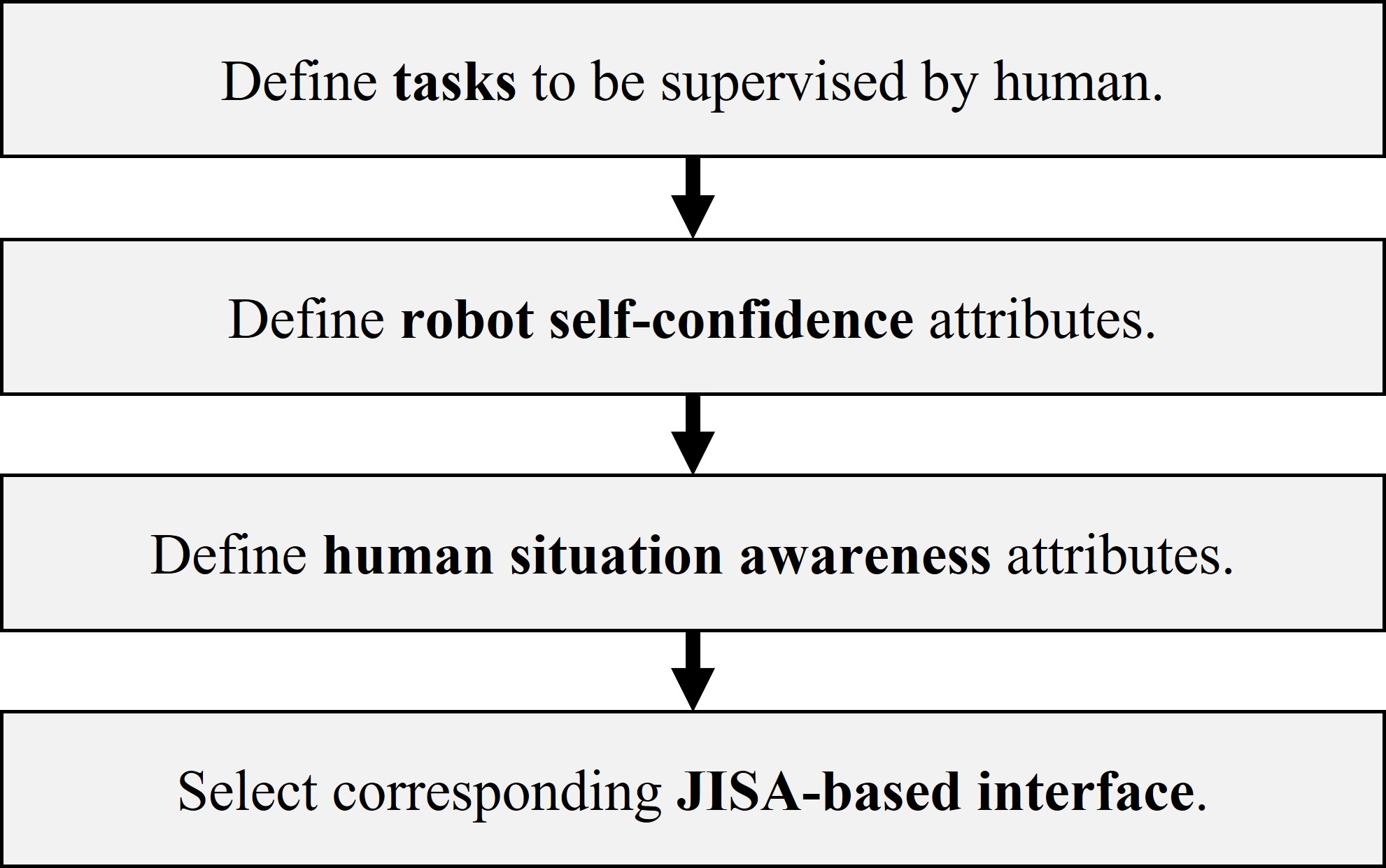}
\centering
\caption{Guideline to apply the proposed JISA framework in autonomous applications with four main modules.}
\label{fig:Meth_flowchart}
\end{figure}

\textbf{1. Define tasks to be supervised by human:} the first step that is necessary to implement JISA is for one to define the tasks in which human supervision is needed based on the challenges and limitations of autonomy in the given system/application. These tasks depend on the robot’s hardware/software, mission type, and the environment in which the robot is operating. 

To understand the challenges and limitations of autonomy in the given system/application, the JISA system designer can run the application in full autonomy mode, and compare the obtained performance and results with those that are expected. Another way is to find common challenges and limitations for similar systems/applications from the literature. After that, the JISA system designer has to define the tasks affected by these challenges and limitations, in which a human could help. For example, some challenges that face an autonomous mobile robot in a SLAM application can be present in obstacles that the robot cannot detect, these challenges affect the mapping task. while other challenges that affect the picking task in a pick-and-place application can be present in objects that are out of the manipulator's reach. 

The output of this step is a list of all the tasks that should be supervised by the human. It is important to mention here that not all tasks in the autonomous application need to be supervised, as some tasks result in better performance when being fully autonomous. 

\textbf{2. Define the robot's self-confidence attributes:} this step is concerned with the SC mechanism used by the robot to reason about when to ask the human for help and what happens if the human does not respond. Depending on the type of task, available internal-states of the robot, and full-autonomy experiments, the robot SC state could be toggled between "confident" and "not-confident" either through (1) detecting certain events (SC-events), and/or through (2) monitoring certain metric/s (SC-metrics) within the supervised task and comparing them to a confidence threshold that is defined experimentally. Examples of events that could be used to set robot SC to "not-confident" are the loss of sensor data, inability to reach the goal, failure in generating motion plans; while examples of monitored metrics are performance, battery level, dispersion of particles in a particle filter, among others.

Depending on the task and how critical it is to get human assistance once requested, a `no-response' strategy should be defined. Examples of such strategy are: wait, request again, wait for defined time before considering the decision approved/declined, put request on hold while proceeding with another task, among others.

For each task to be supervised, the JISA system designer has to determine whether or not it is possible to define events/metrics to be used to set the SC state. Thus, the outputs of this step include (1) SC-based tasks: tasks of which human assistance is requested through SC mechanism, (2) SC metrics and/or SC events, (3) `no response' strategy for each SC-based task, and (4) supervised tasks that human assistance could not be requested through a SC mechanism.

\textbf{3. Define the human situation awareness attributes:} this step is crucial for supervised tasks in which robot SC attributes could not be defined. In such tasks, the assistance has to depend solely on the human SA. Given that on one hand the human is aware of the system’s limitation, and on the other is endowed with superior situational and contextual awareness, s/he can determine when it is best to intervene at a task to attain better results where SC attributes could not be defined. However, the JISA system designer should define the needed tools to enhance human SA and help them better assess the situation. Examples of SA tools include showing certain results of the performed task, and communicating the status of the task planning/execution, to name a few. The outputs of this step are (1) SA-based tasks: tasks in which assistance depends on human SA, (2) tools that would enhance human SA.

\textbf{4. Choose the corresponding JISA-based User interface:} after defining the tasks to be supervised and the SC/SA attributes, a major challenge lies in choosing the proper JISA-based UI that facilitates interaction between the human and the agent. The interface, along with the information to be communicated highly depend on the tasked mission, the hardware/software that are used, and the tasks to be supervised. Thus, the JISA-based UI must be well-designed in order to (1) help both agents communicate assistance requests/responses and human interventions efficiently, (2) maximize the human awareness through applying the SA tools, and (3) govern the type of intervention/assistance so that it does not negatively influence the entire process. JISA-based UIs include, but are not limited to graphical user interfaces, AR interfaces, projected lights, vocals, gestures, etc. The desired output from this step is choosing/designing a JISA-based UI.

\begin{figure}[t]
\centering
\includegraphics[width=3 in]{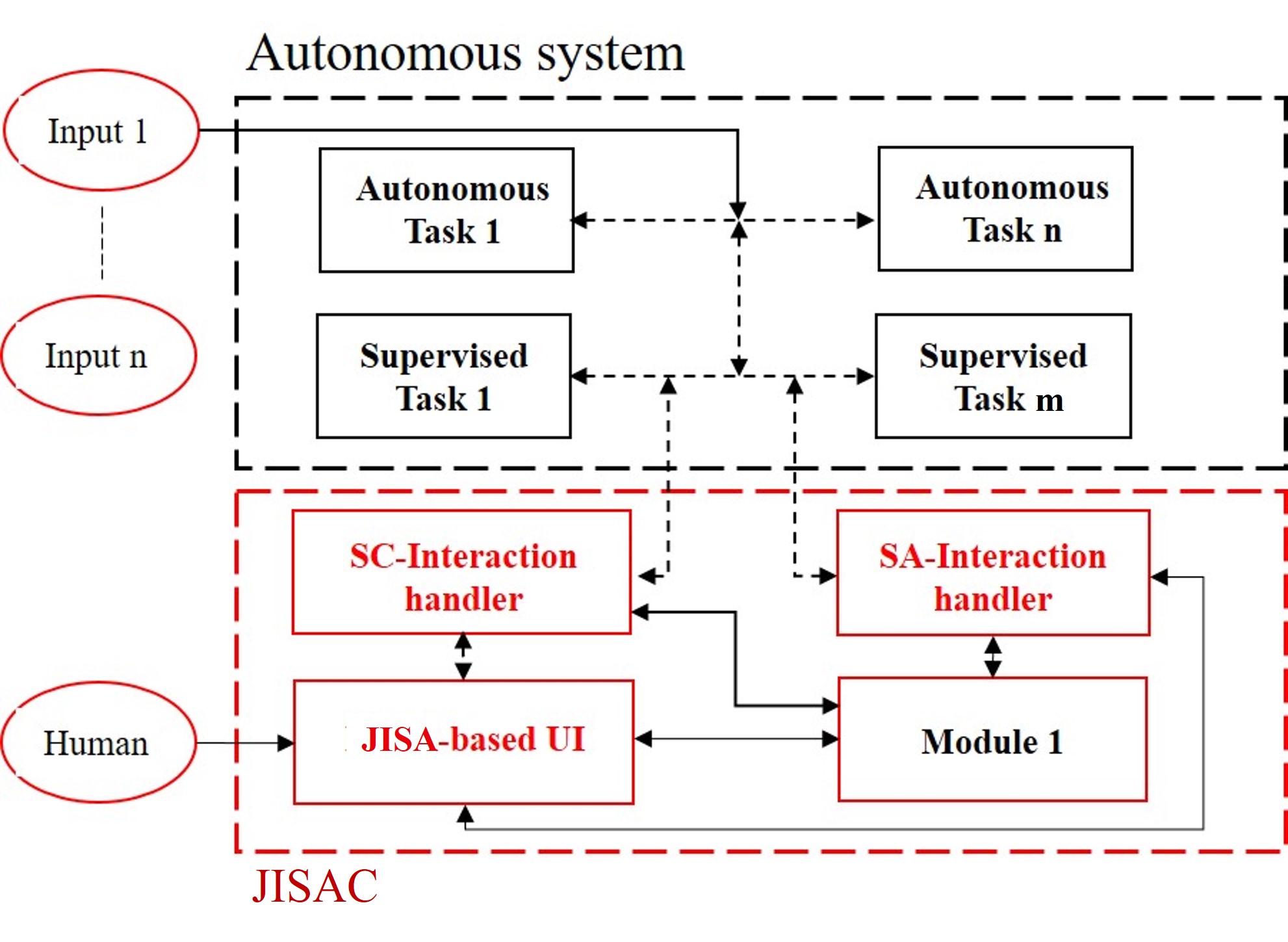}
\caption{Generic diagram of the proposed JISA framework with its main blocks and modules.}
\label{fig:generic_diagram}
\end{figure}

In addition to the discussed guideline, Fig. \ref{fig:generic_diagram} presents a block diagram to help apply JISA in any automated system. This diagram consists of two main blocks: the Autonomous System block and the JISA Coordinator (JISAC) block. 
The autonomous system block contains modules that correspond to the tasks performed by the autonomous agent; as mentioned in the guideline, some of these tasks are supervised, while others are fully autonomous. 
The JISAC block contains three main modules: SC-Interaction handler, SA-Interaction handler, and the JISA-based UI. Interaction based on robot SC is handled through the SC-Interaction handler. This module monitors the SC-events/metrics defined through the `robot SC attributes' and sends assistance requests to the human supervisor through the JISA-based UI. When the human responds to the request, SC-Interaction handler performs the needed operations and sends back the results to the corresponding supervised task. If no response from the human is received, the SC-Interaction handler applies the no-response strategy along with the needed operations.

Since an operator monitors the system through the SA tools provided in the JISA-based UI, he can intervene at any time based on SA. The SA-Interaction handler receives the human input/intervention, applies the needed operations, and sends the results to the corresponding supervised task. It is important to mention that both SC- and SA-Interaction handlers can be connected to several supervised tasks; and thus could handle different types of interactions and operations. Moreover, the JISAC block could contain other modules that handle some operations not directly related to the SC- SA- interaction handlers.  
To better describe how JISA should be implemented, in the following two sections we apply the suggested guideline attributes and block-diagram to two autonomous tasks, including collaborative SLAM and automated puzzle solving. 

\section{JISA in Collaborative SLAM}
\label{sec:JISA_SLAM}
This section constitutes the first of two case studies in which JISA is applied. Although the core of this case study is presented in \cite{aug_mapping, aug_SLAM, coll_SLAM}, the works are re-framed (as POC) in the context of JISA. This is done to demonstrate the applicability of JISA in different robotic applications, and to serve as an example of how to apply the framework in a SLAM application. Thus, this section is considered as an\textit{ evolution} and \textit{generalization} of the approach presented in \cite{aug_mapping, aug_SLAM, coll_SLAM}, where we intend to show how the JISA guideline is followed and the framework is applied, rather than presenting `JISA in Collaborative SLAM' as a standalone contribution and generating new results.

The idea of collaborative SLAM is not new. An online pose-graph 3D SLAM for multiple mobile robots was proposed in \cite{c_m3}. This method utilizes a master-agent approach to perform pose-graph optimization based on odometry and scan matching factors received from the robots. Since a centralized approach is used for map merging, SLAM estimates in this approach may be affected by major errors in the case of connectivity loss. Some researches proposed to localize an agent in the map built by another agent \cite{c_m8,c_m4}. The problem with such approaches is that one agent fails to localize when navigating in an area that is not yet mapped by the second agent. 

Developing HRI systems where the efficiency and perception of humans aid the robot in SLAM has also gained some interest. For instance, \cite{sl19} showed that augmenting the map produced by a robot with human-labelled semantic information increases the total map accuracy. However, this method required post processing of these maps. \cite{c_m18} proposed to correct scan alignments of 3D scanners through applying virtual forces by a human via a GUI. since this method lacks localization capability, it cannot produce large accurate maps. Sprute \textit{et al.} \cite{Joyce_Sprute}  proposed a system where a robot performs SLAM to map an area then the user can preview this map augmented on the environment through a tablet and can manually define virtual navigation-forbidden areas. Although these systems proved superior over fully autonomous methods by being more efficient in increasing map accuracy, they totally depend on human awareness and do not consider asking the human for help in case of challenging situations.

In our implementation of JISA in collaborative SLAM, the effects of delays and short communication losses are minimized since each agent runs SLAM on its own. Moreover, the proposed system utilizes one agent to correct mapping errors of another agent under the supervision of a human operator, who can also intervene based on the agents' requests or his judgment. The AR-HMD is also used to (1) visualize the map created by a robot performing SLAM aligned on the physical environment, (2) evaluate the map correctness, and (3) edit the map in real-time through intuitive gestures.

The proposed application of JISA in collaborative SLAM allows three co-located heterogeneous types of agents (robot, AR head mount device AR-HMD, and human) to contribute to the SLAM process. Applying JISA here allows for real-time pose and map correction. Whenever an agent’s self-confidence drops, it asks the human supervisor to approve/correct its pose estimation. In addition, the human can view the map being built superposed on the real environment through an AR interface that was specifically developed for this purpose, and apply edits to this map in real-time based on his/her SA. Moreover, when the robot is unable to reach a certain navigation goal, it prompts the human to assist in mapping the area corresponding to this goal; the human here is free to choose between mapping the area through editing the map him/herself or through activating the AR-HMD collaborative mapping feature. In the latter case, the AR-HMD contributes to the map building process by adding obstacles that are not detected by the robot and/or map areas that the robot cannot traverse. 

\begin{figure}[t]
\includegraphics[width=3.2 in]{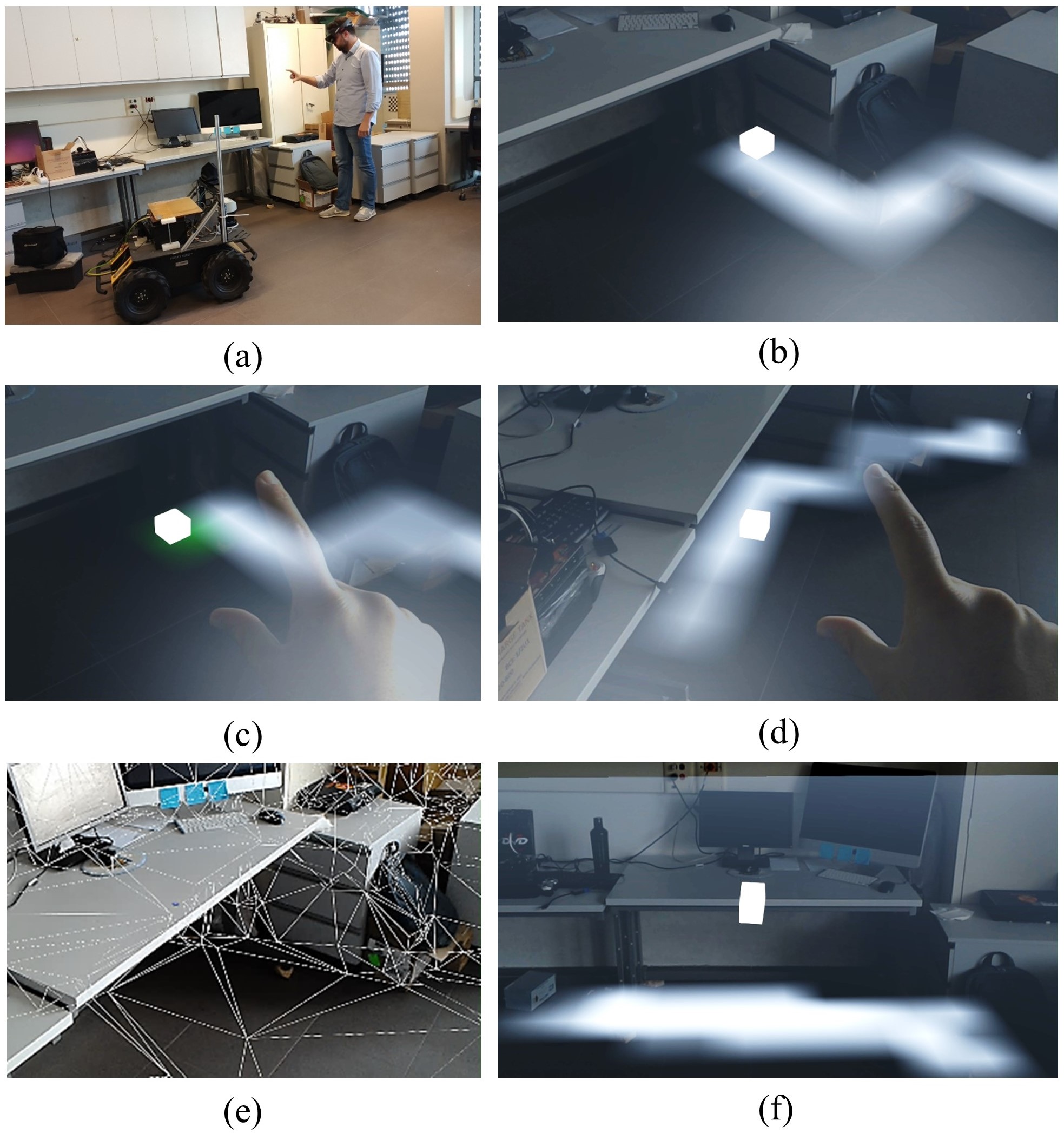}
\centering
\caption{Sample demonstration of the proposed JISA in a collaborative SLAM system. (a) Robot alongside a human operator wearing an AR-HMD, (b) the augmented map with white lines representing the occupied cells, (c) user is adding occupied cells, (d) the desk boundary added by the user, (e) the 3D mesh created by the AR-HMD, and (f) projection of the table in the robot map \cite{coll_SLAM}.}
\label{fig:sample_m}
\end{figure}

Fig. \ref{fig:sample_m} presents a sample demonstration of the proposed system, where in Fig. \ref{fig:sample_m}a we show an operator wearing a HoloLens with the nearby robot performing SLAM. Fig. \ref{fig:sample_m}b shows how the occupancy grid, produced by the robot, is rendered in the user’s view through a HoloLens, where the table, which is higher than the robot’s LiDAR plane, is not represented in the map. Fig. \ref{fig:sample_m}c demonstrates the addition of occupied cells representing the table (in white), and Fig. \ref{fig:sample_m}d shows how the boundaries of the table are added by the user. Fig. \ref{fig:sample_m}e shows how the HoloLens detects the table and visualize its respective 3D mesh. Finally, Fig. \ref{fig:sample_m}f shows the corresponding projection of the table merged in the occupancy map.
\subsection{System Overview}
This section describes how the proposed JISA framework is applied to collaborative grid-based SLAM, following the methodology presented in Section (\ref{sec:meth}). Table \ref{table:JISA_SLAM} summarizes the inputs form the SLAM problem, which are used to satisfy the JISA framework.
\begin{table*}[t]
    \centering
    \caption{JISA attributes in collaborative SLAM applications.}
    \begin{tabularx}{6.85 in}{| >{\centering\arraybackslash}X| >{\centering\arraybackslash}X| >{\centering\arraybackslash}X| >{\centering\arraybackslash}X| >{\centering\arraybackslash}X|}
    \hline
    \multicolumn{2}{|c|}{\textbf{Tasks to be supervised}} & \multicolumn{3}{c|}{\textbf{Attributes}} \\ 
    \hline
    SC-based tasks & SA-based tasks & SC Attributes& SA Attributes & JISA-based UI\\ 
    \hline
    \begin{itemize}[leftmargin=*]
        \item Robot pose estimation
        \item AR-HMD pose estimation
        \item Ability to reach navigation goal
    \end{itemize}
    &
    \begin{itemize}[leftmargin=*]
        \item Map building by robot
        \item Map building by AR-HMD
        \item Global map merging
    \end{itemize}
    &
    \begin{itemize}[leftmargin=*]
        \item Confidence in pose 
        \item Confidence in reaching navigation goals
        \item SC-metric: $N_{eff}$
        \item SC-event: HoloLens tracking feature
    \end{itemize}
    &
    \begin{itemize}[leftmargin=*]
        \item Maps quality 
        \item SA-tools: Augment the map on physical world
    \end{itemize}  
    &
    \begin{itemize}[leftmargin=*]
        \item AR interface
        \item 2D GUI
    \end{itemize}      
    \\
    \hline
    \end{tabularx}
    \label{table:JISA_SLAM}
\end{table*}

\textbf{1. Tasks to be supervised by the human:} the main challenges facing SLAM are inaccurate map representation and inaccurate localization \cite{aug_SLAM}. In SLAM, mapping and localization are performed simultaneously, and thus increased accuracy in any of the two tasks enhances the accuracy of the other, and vice versa. In this system, the human should be able to supervise and intervene in the following tasks: (1) map building by the robot, (2) map building by the AR-HMD, (3) robot pose estimation, (4) AR-HMD pose estimation, and (5) global map merging.

\textbf{2. Robot SC attributes:} as discussed in \cite{aug_SLAM}, the robot's SC in the pose estimation is calculated through the effective sample size ($N_{eff}$) metric which reflects the dispersion of the particles' importance weights. This metric serves as an estimation of how well the particles represent the true pose of the robot; thus, $N_{eff}$ is employed as SC-metric to reason about when the robot should prompt the human for assistance in localizing itself. In this system, the robot stops navigating and prompts the human to approve or correct the estimated pose whenever $N_{eff}$ drops below a threshold that was obtained experimentally. As for the HoloLens, the human supervisor is requested to help in the pose estimation based on a built-in feature that flags the HoloLens localization error event. Since any error in localization affects the whole map accuracy, the system does not resume its operation unless human response is received. Moreover, when the robot is given a navigation goal that it is not able to find a valid path to, it prompts the human supervisor to help in mapping the area corresponding to this goal. For this task, the robot proceeds into its next goal even if the human does not respond to the request directly. 

\textbf{3. Human SA attributes:} since SC attributes could not be defined in the map building task (for both robot and HoloLens), the applied JISA system benefits from the human perception and SA in assessing the quality of maps built by both agents. To help the human supervisor compare the map built by the agents with the real physical environment and apply edits, the SA tool proposed is augmenting the map on the physical world. The human is allowed to edit the robot's map by toggling the status of cells between `occupied' and `free'. Moreover, the human can choose to activate the collaboration feature to assist the robot in mapping whenever he finds it necessary. 

\textbf{4. JISA-based user interface:} to maximize the human's SA and provide an intuitive way of interaction, an augmented reality interface (ARI) running on HoloLens is developed. Through this interface, humans can see the augmented map superposed over the real environment and can interact with it through their hands. Moreover, and upon request, humans can influence the robot's pose estimation through a GUI.

We are applying our proposed JISA framework on top of OpenSLAM Gmapping \cite{sl33}: a SLAM technique that applies Rao-Blackwellized particle filter (RBPF) to build grid maps from laser range measurements and odometry data \cite{sl27}. An RBPF consists of $N$ particles in set $R_{t}$ where each particle ${R_{t}^{(i)}} = (\zeta_{t}^{(i)}, \eta_{t}^{(i)}, w_{t}^{(i)})$ is presented by a proposed map $\eta_t^{(i)}$, pose $\zeta_t^{(i)}$ inside this map, and an importance weight  $w_t^{(i)}$. OpenSLAM can be summarized by five main steps \cite{sl27}: Pose Propagation, Scan Matching, Importance Weighting, Adaptive Resampling, and Map Update.

Fig. \ref{fig:SLAMmeth_flowchart} presents the flowchart of JISA framework in collaborative SLAM. Description of modules in OpenSLAM gmapping block can be found in \cite{aug_SLAM}, while description of modules in the JISAC block are summarized below: 
\begin{figure}[b]
\centering
\includegraphics[width=3.2 in]{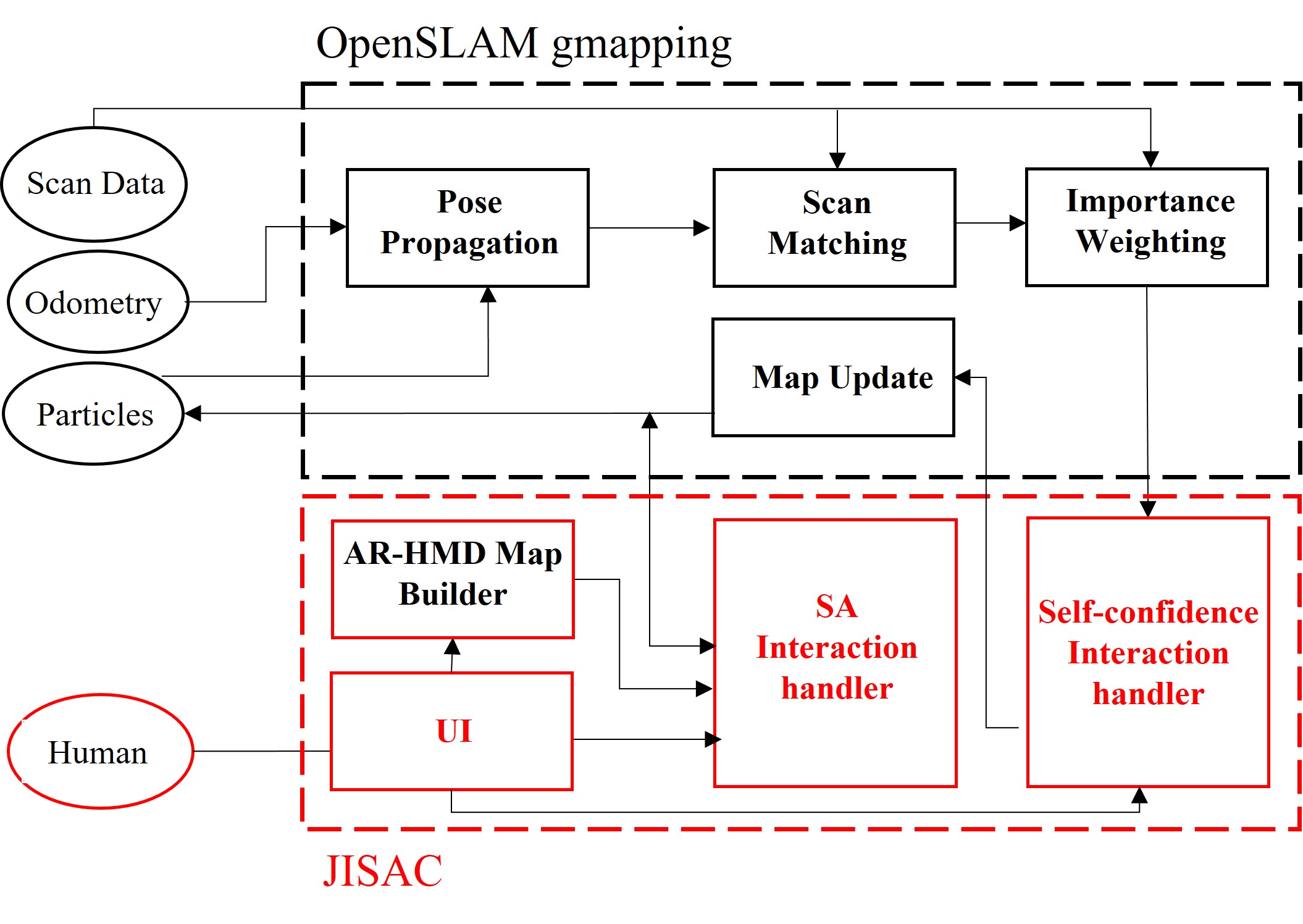}
\caption{The proposed JISA-based SLAM system flowchart: blocks in black are adopted from OpenSLAM and blocks in red represent the proposed JISAC.}
\label{fig:SLAMmeth_flowchart}
\end{figure}
\subsubsection{SC-Interaction handler}
This module represents the mechanism applied to check the self-confidence of the robot and HoloLens and prompts the human supervisor for help through the JISA-based UI. When the HoloLens SC-event is triggered, this module prompt the human supervisor to re-localize the HoloLens. As for the robot, the SC is obtained based on the $N_{eff}$ metric given in equation (\ref{eq:Neff}). If $N_{eff} > r_{th}$, where $r_{th}$ is a confidence threshold obtained through empirical testing, no human assistance is requested; thus next steps are ignored and $R_t$ is passed as-is to the OpenSLAM gmapping block.
\begin{equation}\label{eq:Neff}
        N_{eff} = \frac{1}{\sum_{i=1}^{N}(\tilde{w}^{(i)})^2 },
\end{equation}
If $N_{eff} < r_{th}$, the human is requested to approve or correct the pose estimation through the GUI. When the new human-corrected pose is acquired, an augmented pose $\acute{a}_t^i \sim \mathcal{N}(\mu,\Sigma)$ is calculated for each particle. This Gaussian approximation is calculated since the human correction itself may include errors. After that, we perform \textit{scan matching} followed by \textit{importance weighting}, and finally we update all the particles' poses in $R_t$. 
\subsubsection{SA-Interaction handler}\label{sec:SA-coll}
the SA-Interaction handler is responsible for fusing the human edits and the AR-HMD map with the robot's map to produce a global map that is used by all agents. This module contains two functions:\\
\begin{enumerate}
\item{AR Map Merger:} this function produces a `merged map' to be viewed by the human operator. The merged map is constructed by comparing the human-edited cells and the updated cells from the AR-HMD Map Builder module to the corresponding cells of the robot’s map. The function always prioritizes human edits over the robot and the AR-HMD maps, and it prioritizes Occupied status over Free whenever the cell status is not consistent between the robot map and the AR-HMD map.
\item{Global Map Merger:} this function fuses the merged map with the most up-to-date robot map to produce a global map that is used by all agents. This ensures that whenever the robot has gained confidence about the state of a cell, its knowledge is propagated to the other agents. To avoid collision between the robot and undetected obstacles, cells that are set to be Occupied by the AR Map Merger, but found to be free by the robot, are set as Occupied in this module.
\end{enumerate}
\subsubsection{AR-HMD Map Builder}
This module produces an AR-HMD occupancy grid map in two steps: (1) \textit{ray-casting} to detect the nearest obstacles to the AR-HMD within a defined window, and (2) \textit{map updating} to build and update the AR-HMD occupancy grid map based on the ray casting output as discussed in \cite{coll_SLAM}. The produced map shares the same size, resolution, and reference frame with the robot’s map. The AR-HMD device is assumed to be able to create a 3D mesh of the environment, and the relative transformation between the robot’s map frame and the AR-HMD frame is known.
\subsection{Experiments and Results}
\begin{figure}[b]
\centering
\includegraphics[width=2.8 in]{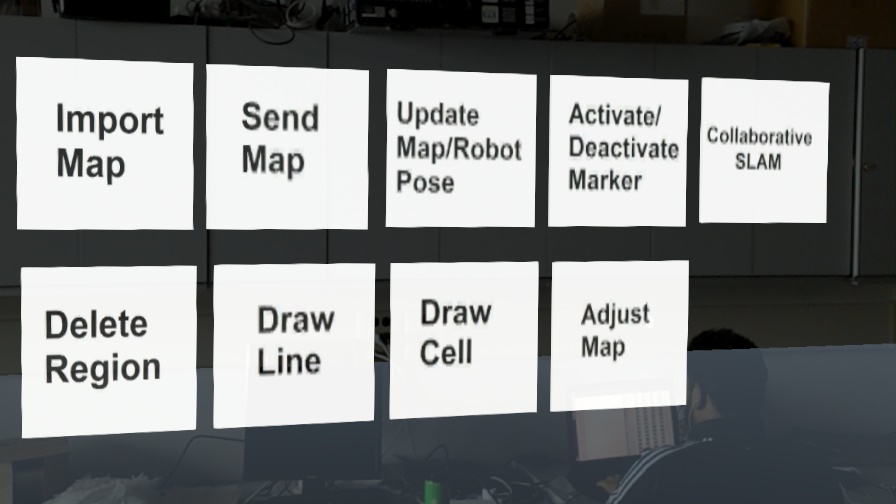}
\caption{Graphical user interface on the HoloLens, showing the menu items that are displayed to the left of the user \cite{coll_SLAM}.}
\label{fig:menu}
\end{figure}

The proposed JISA framework was implemented using Robotics Operating System (ROS) Indigo distro and Unity3D  2018.4f1. 
The interaction menu developed in the JISA-based UI is shown in Fig. \ref{fig:menu}. This menu allows the user to import/send augmented maps, initialize/re-initialize a map pose, manually adjust the pose of augmented map in the physical world, activate/deactivate the AR-HMD map builder, and choose to edit the map through: toggling the status of individual cells, drawing lines, and deleting regions.

\begin{figure}[t]
\centering
\includegraphics[width=3 in]{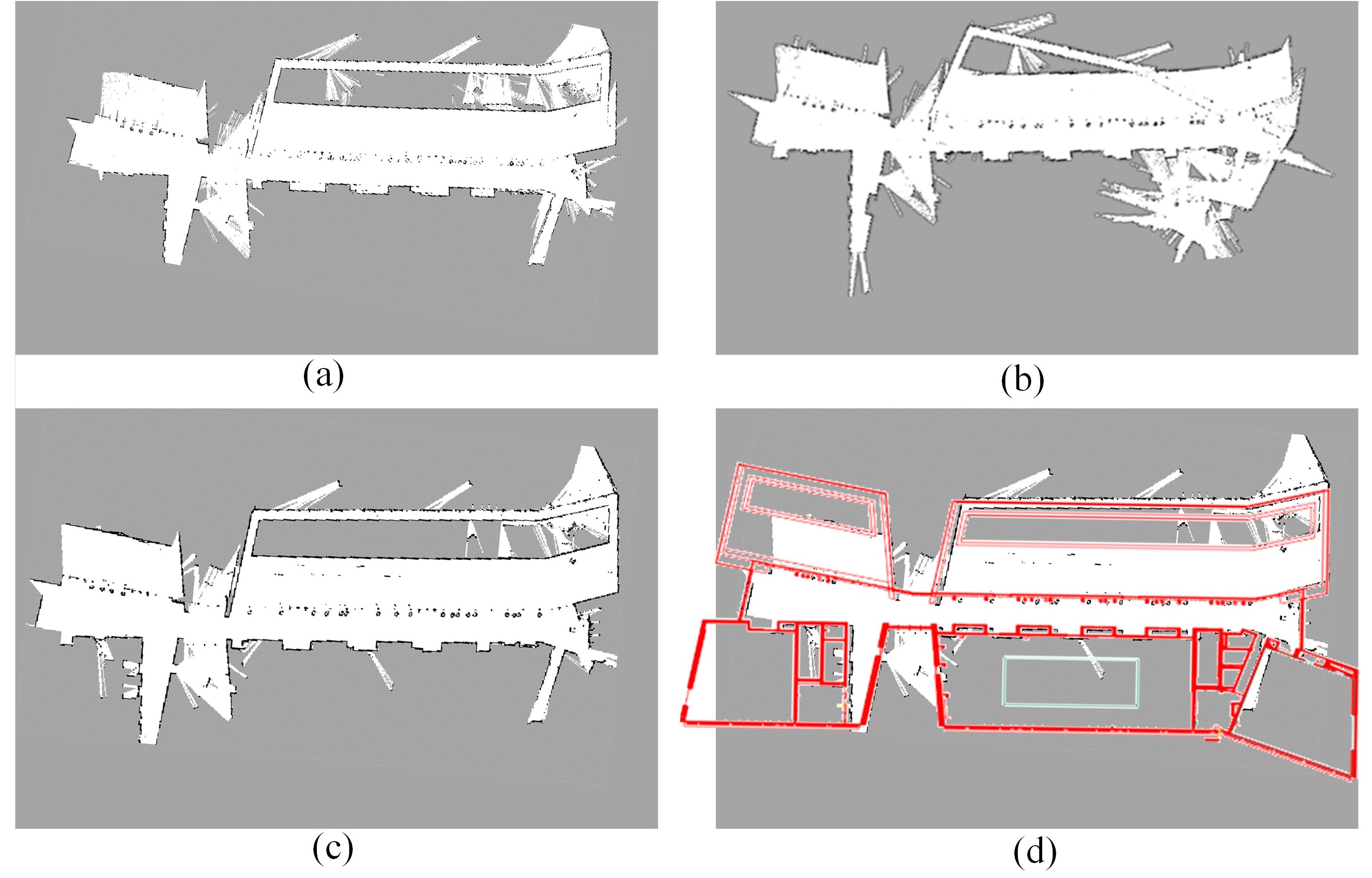}
\caption{Maps obtained using Set C (high laser range) \cite{aug_SLAM}.}
\label{fig:Maps_SetC}
\end{figure}

Testing for this system was conducted in Irani Oxy Engineering Complex (IOEC) building at the American University of Beirut (AUB). The testing areas were carefully staged in a way to introduce challenges that affect the performance of SLAM (\textit{i.e.}, low semantic features, glass walls, object higher and others lower than the LiDAR's detection plane). Clearpath Husky A200 Explorer robot and HoloLens were utilized during the experimentation process. 
The first batch of experiments was performed to evaluate correcting the robot’s pose on the overall accuracy of the resulting SLAM maps. In this batch, the operator had to tele-operate the robot, following a defined trajectory, for a distance of about $170m$. Table \ref{table:sets} presents the parameters used in the different sets of experiments, where each set of tests was run three times, resulting in 54 tests. 
Furthermore, using the proposed JISA framework, three experiments are conducted using parameters that resulted in the worst map representation for each set.
\begin{table}[t]
\centering
\scriptsize
\caption{Experimentation sets and their parameters \cite{aug_SLAM}.}
\begin{tabular}{|c|c|c|c|c|c|c|c|c|c|} 
 \hline
 Set &\multicolumn{3}{|c|}{A} &\multicolumn{3}{|c|}{B} &\multicolumn{3}{|c|}{C}
\\
\hline
 Lighting &\multicolumn{3}{|c|}{Day, Night} &\multicolumn{3}{|c|}{Day,  Night} &\multicolumn{3}{|c|}{Day,  Night}
\\
\hline
 Laser range &\multicolumn{3}{|c|}{5.5-6.1 m} &\multicolumn{3}{|c|}{7.5-8.1 m} &\multicolumn{3}{|c|}{9.5-10.1 m}
\\
\hline
Particles &5 &50 &100 &5 &50 &100 &5 &50 &100
\\
\hline
 \end{tabular}

\label{table:sets}
\end{table}
\begin{figure}[b]
\centering
\includegraphics[width=3 in]{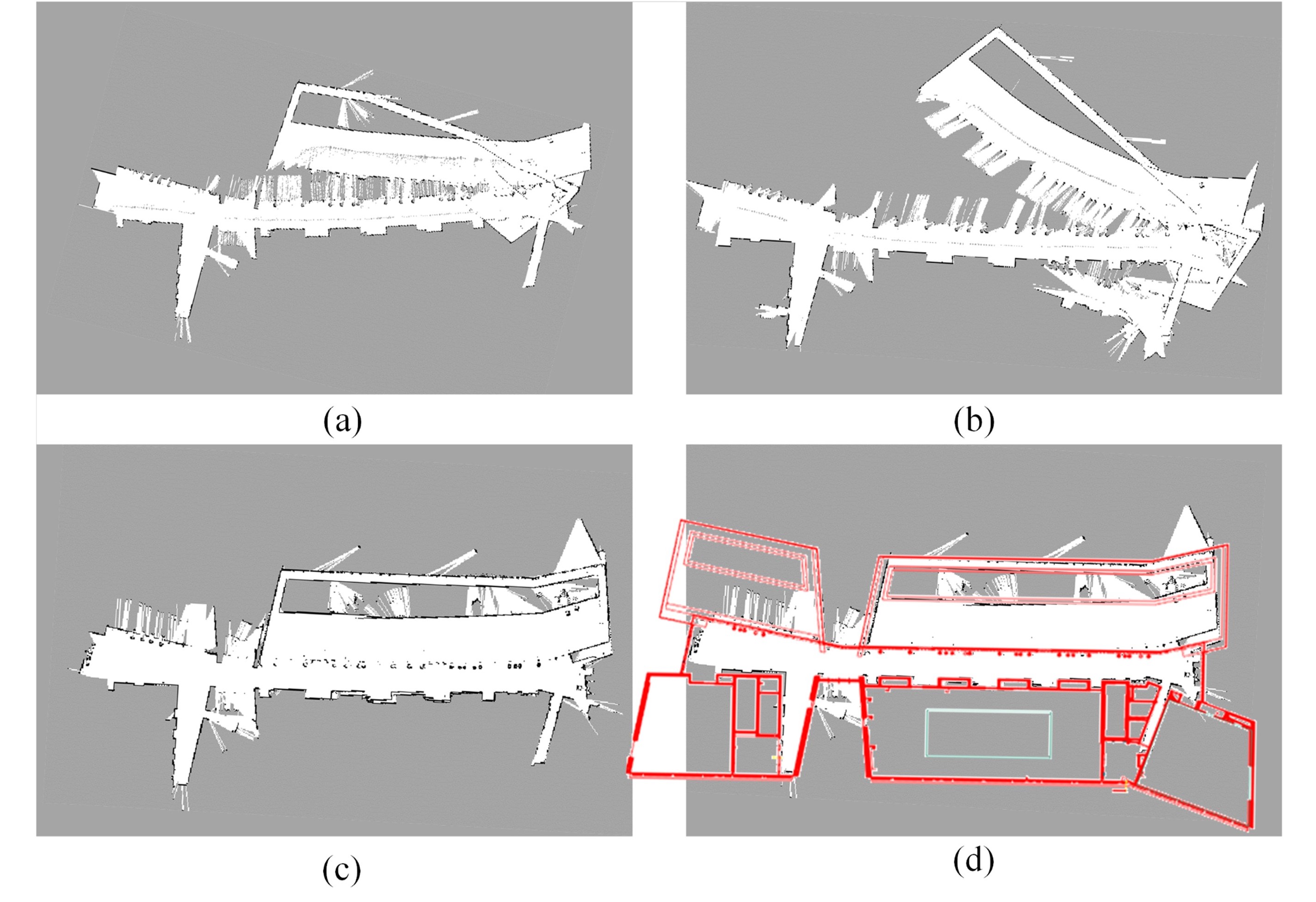}
\caption{Maps obtained using Set A (low laser range) \cite{aug_SLAM}.  }
\label{fig:Maps_SetA}
\end{figure}
Experiments show that the accuracy of maps increases with the increase in the laser range and number of particles. However, this improvement comes at a higher computational cost.The best maps obtained through full autonomy using parameters in Sets A and C are shown in Fig. \ref{fig:Maps_SetA}a and Fig. \ref{fig:Maps_SetC}a, while the worst maps are shown in Fig. \ref{fig:Maps_SetA}b and Fig. \ref{fig:Maps_SetC}b respectively. Fig. \ref{fig:Maps_SetA}c and Fig. \ref{fig:Maps_SetC}c show the results of the maps using the proposed JISA framework, which are overlaid by the blueprints of the area in Fig. \ref{fig:Maps_SetA}d and Fig. \ref{fig:Maps_SetC}d. The conducted experiments demonstrate how implementing JISA to request human assistance in pose estimation can result in higher quality maps, especially in areas that are considered challenging. 
Tests using the JISA framework took $35\% - 50\%$ less time than those performed with OpenSLAM gmapping, since there was no need to revisit areas and check for loop closures. Throughout JISA tests, the operator needed to change the robot's pose in only $9.5\%$ of the times where the SC mechanism requested assistance. 

We define \textit{success rate} as the number of useful maps, which represent the mapped environment correctly with no areas overlaying on top of each other, over the total number of maps obtained throughout the runs. Experiments using OpenSLAM gmapping resulted in a $57.4\%$ success rate. Through the JISA framework, all produced maps had no overlaying areas, resulting in $100\%$ success rate. Moreover, maps produced through JISA did not require any post processing since all map edits where done throughout the runs. 

\begin{figure}[t]
\centering
\includegraphics[width=1.5 in ]{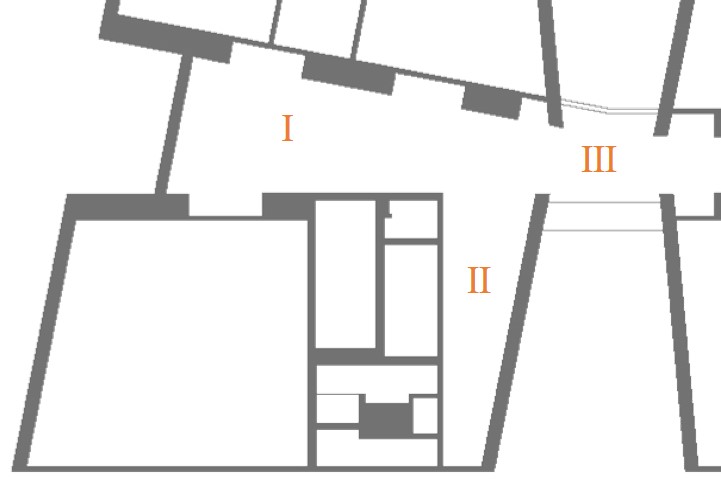}
\caption{Blueprint of the testing arena and photos of the experimental setup. \textbf{Region \RN{1}} contains obstacles that could not be detected by the LiDAR. \textbf{Region \RN{2}} is a corridor with poorly textures walls, and \textbf{Region \RN{3}} has two glass walls facing each other.}
\label{fig:mSLAM_exp}
\end{figure}

\begin{figure}[!t]
\centering
\includegraphics[width=3.2 in]{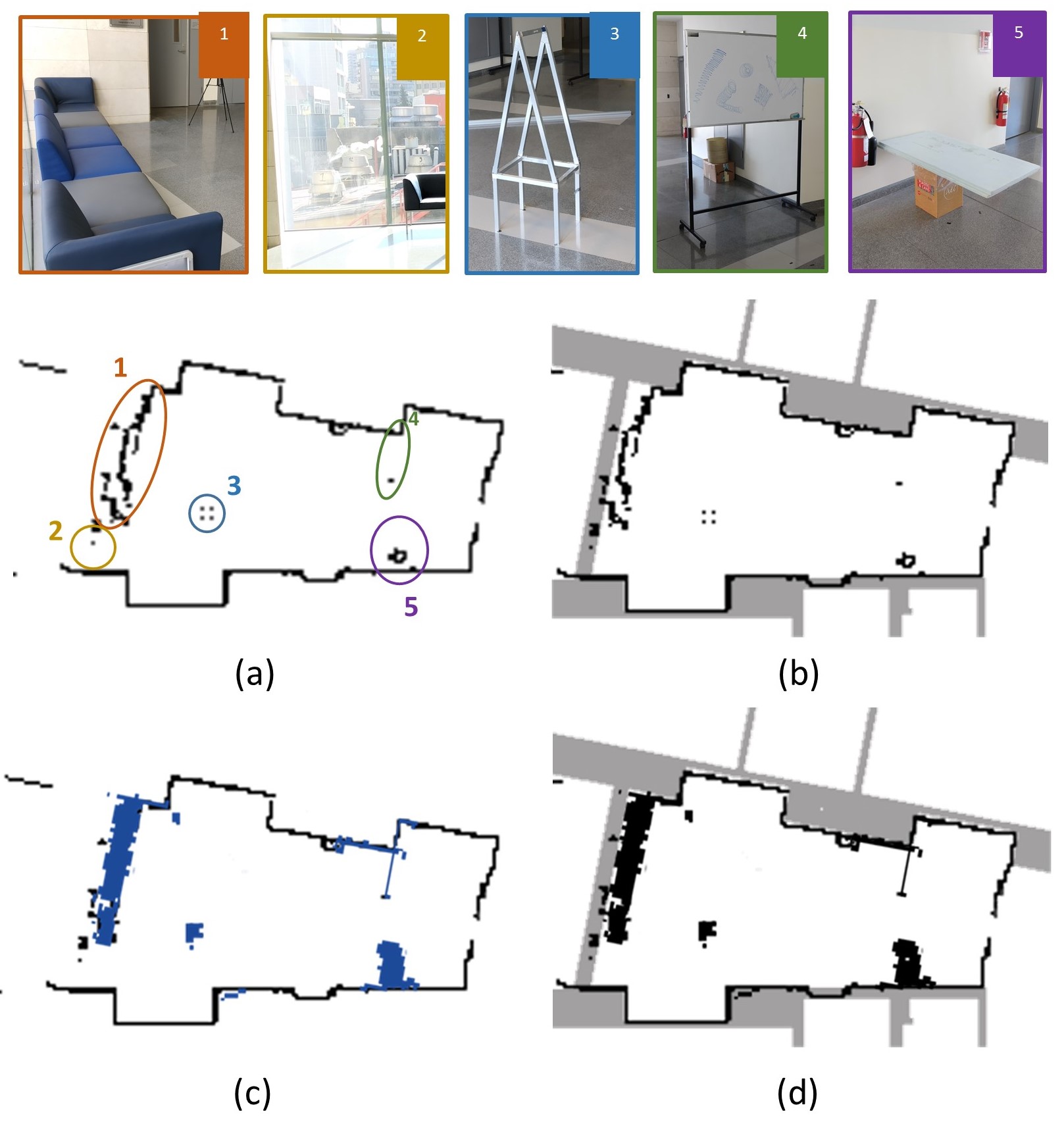}
\caption{(a) The map obtained by the robot where it was not able to detect the objects numbered 1 to 5, which is overlaid on the blueprint in (b). (c) The automatic real-time updates performed by the HoloLens on the augmented map (blue color) merged with the robot’s map, which is overlaid on the blueprint in (d) \cite{coll_SLAM}.}
\label{fig:mSLAM_maps}
\end{figure}

Another experiment was done to assess the implementation of JISA framework in collaborative SLAM. Fig. \ref{fig:mSLAM_exp} shows a blueprint of the arena. Region \RN{1} is separated from Region \RN{2} by an exit that is too narrow for the robot to pass through. 
The robot was tele-operated to map Region \RN{1}. The resultant map is shown in Fig. \ref{fig:mSLAM_maps}a, and it is overlaid on the blueprint in Fig. \ref{fig:mSLAM_maps}b. The robot failed to correctly map some objects such as those numbered from 1 to 5 in Fig. \ref{fig:mSLAM_maps}. The operator had to activate the \textit{AR-HMD Map Builder} module and walk around the un-mapped objects to allow the HoloLens to improve the map. Fig. \ref{fig:mSLAM_maps}c shows the updates performed by the HoloLens, in real-time, in blue color, and Fig. \ref{fig:mSLAM_maps}d shows the merged map overlaid on the blueprint. 

\begin{figure}[ht]
\centering
\includegraphics[width=3.2 in]{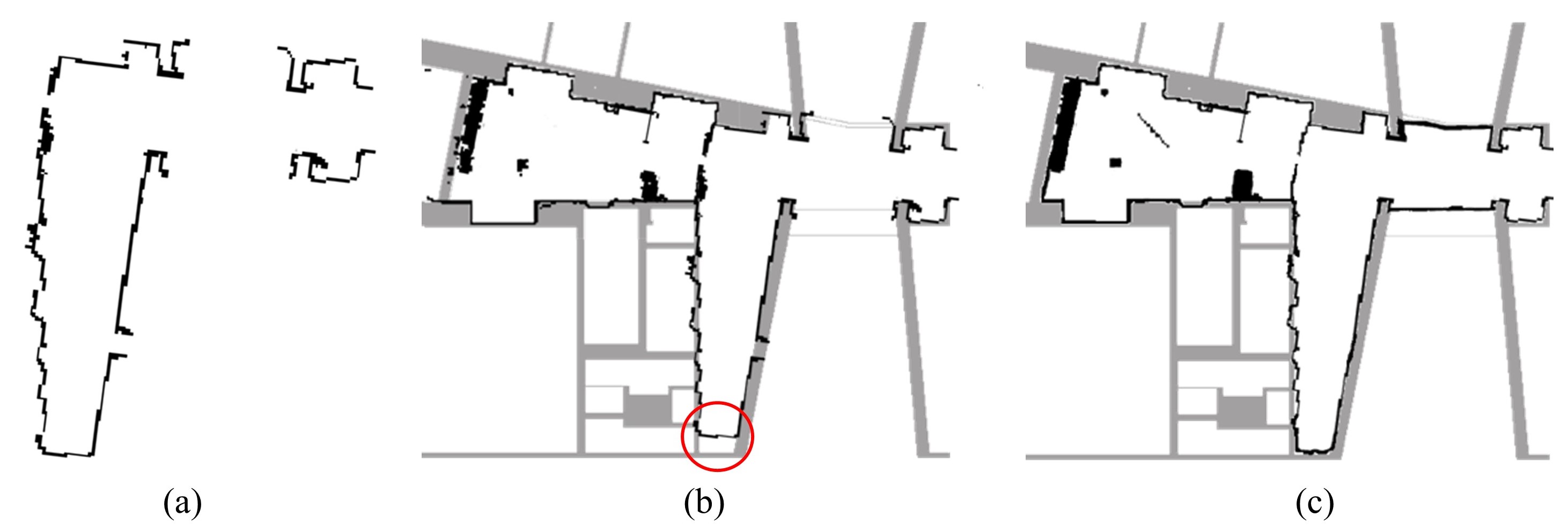}
\caption{(a) Map produced by the AR-HMD Map Builder module in \textbf{Region \RN{2}}. (b) the merged map of both \textbf{Region \RN{2}} and \textbf{Region \RN{3}}; the red circle shows a mapping error that occurred because the HoloLens failed to perform tracking in this location. (c) the final global map after performing human edits to correct all errors \cite{coll_SLAM}.}
\label{fig:mSLAM_ARmap}
\end{figure}

Since the robot cannot get out of Region \RN{1}, the operator walked around regions \RN{2} and \RN{3} while activating the \textit{AR-HMD Map Builder}. Fig. \ref{fig:mSLAM_ARmap}a and Fig. \ref{fig:mSLAM_ARmap}b shows the resulting HoloLens map and the merged map respectively. The HoloLens failed to perform tracking in a feature-deprived location (see red circle in Fig. \ref{fig:mSLAM_ARmap}b), and the glass walls were not detected by the HoloLens. To account for these errors, the human operator performed manual edits while walking around. The final corrected global map is shown in Fig. \ref{fig:mSLAM_ARmap}c. 

The entire experiment took 13 minutes and 50 seconds. To assess this performance, another experiment was conducted in which the robot was tele-operated to map all three regions, and the traditional method of tape-measuring and then editing offline using GIMP software was used. This traditional mapping and post-processing method required around 46 minutes. Thus, the proposed JISA framework \textbf{eliminated post-processing} of the map and needed approximately 70\% less time than the traditional method. 

Through the above experiments, the efficiency of the proposed JISA framework is demonstrated in collaborative SLAM systems by producing more accurate maps in significantly less time. 

\section{JISA in Automated Puzzle Reconstruction}
\label{sec:JISA_puzzle}
The Jigsaw puzzle is a problem that requires reconstructing an image from non-overlapping pieces. This `game' has been around since the 18th century; however, the first computational approach to solve a puzzle was introduced in 1964 \cite{c_p13}. In addition to being an interesting and challenging game, researchers have applied puzzle solving techniques in different applications such as DNA/RNA modeling \cite{c_p15}, speech descrambling \cite{c_p14}, reassembling archaeological artifacts \cite{c_p17}, and reconstructing shredded documents, paints, and photos \cite{c_p18, c_p19, c_p21}. 

Jigsaw puzzles are generally solved depending on functions of colors, textures, shapes, and possible inter-piece correlations. In recent years, research into solving jigsaw puzzles has focused on image puzzles with square non-overlapping pieces; thus, solving these puzzles has to rely on the image information only. The difficulty of such puzzles varies depending on the type of pieces that they consist of. These types can be categorized into pieces with (1) unknown orientations, (2) unknown locations, and (3) unknown locations and orientations, which is the hardest to solve.

The first algorithm to solve jigsaw puzzles was proposed by Freeman and Gardner \cite{c_p35} in 1964 where the focus was on the shape information in puzzle pieces. Using both edge shape and color information to influence adjoining puzzle pieces was first introduced in \cite{c_p8}, where the color similarity along matching edges of two pieces was compared to decide on the reconstruction. Other approaches that were proposed for such puzzles relied on isthmus global critical points \cite{c_p38}, dividing puzzle pieces to groups of most likely `interconnected' pieces \cite{c_p11}, applying shape classification algorithms \cite{c_p382}, among others.

Taking the challenge a step further, several researchers proposed algorithms to solve jigsaw puzzles of square pieces with unknown locations and orientations, thus the reconstruction of pieces depends on image information alone. This type of puzzles was proven to be an NP-complete problem \cite{c_p1}, meaning that formulating it as an optimization problem with a global energy function is very hard to achieve \cite{c_p7}. \cite{c_p1} proposed to solve square puzzles through a graphical model and probabilistic approach that uses Markov Random Field and belief propagation. However, this method required having anchor pieces given as prior knowledge to the system. Gallagher \cite{c_p5} also proposed a graphical model that solves puzzles through tree-based reassembly algorithm by introducing the Mahalanobis Gradient Compatibility (MGC) metric. This metric measures the local gradient near the puzzle piece edge. Zanoci and Andress \cite{c_p13} extended the work in \cite{c_p5} and formulated the puzzle problem as a search for minimum spanning tree (MST) in a graph. Moreover, they used a Disjoint Set Forest (DSF) data structure to guarantee fast reconstruction. One of the common limitations of these approaches is that the algorithm cannot change the position of a piece that has been misplaced early on in the reconstruction process. Such mistakes lead to very poor overall performance of the algorithm in many cases.

In our proposed work, we adopt the methods applied in \cite{c_p5} and \cite{c_p13} in forming the problem as a search for MST, applying DSF, and using the MGC metric for compatibility. 
The proposed JISA framework addresses the challenges presented in solving image puzzles with square pieces of unknown locations and orientations, using the greedy approach. Through a custom-developed GUI, the human supervisor monitors the puzzle reconstruction results in real-time, intervenes based on his/her SA, and receives assistance requests from the agent when its self-confidence drops. In addition, the GUI communicates some internal states of the agent such as the reconstruction completion percentage, pieces that are being merged, and the location of the merged pieces in the reconstruction results. We assume that based on the SA, imagination, and cognitive power of a human, s/he can look at puzzle pieces or clusters and be able to detect errors, intervene, and asses matching options, thanks to their ability to see `the big picture.' 

\begin{table*}[t]
    \centering
    \caption{JISA attributes in Automated Puzzle Reconstruction.}
    \begin{tabularx}{6.85 in}{| >{\centering\arraybackslash}X| >{\centering\arraybackslash}X| >{\centering\arraybackslash}X| >{\centering\arraybackslash}X| >{\centering\arraybackslash}X|}
    \hline
    \multicolumn{2}{|c|}{\textbf{Tasks to be supervised}} & \multicolumn{3}{c|}{\textbf{Attributes}} \\ 
    \hline
    SC-based tasks & SA-based tasks & SC Attributes& SA Attributes & JISA-based UI\\ 
    \hline
    \begin{itemize}[leftmargin=*]
        \item Matching of poorly textured pieces
    \end{itemize}
    & 
    \begin{itemize}[leftmargin=*]
        \item Global puzzle reconstruction
        \item Trim frame location
    \end{itemize} 
    &
    \begin{itemize}[leftmargin=*]
        \item Confidence in local pieces merging
        \item SC-metric: Piece entropy
    \end{itemize}
    &
    \begin{itemize}[leftmargin=*]
        \item Global puzzle consistency
        \item SA-tools:
        \begin{itemize}
            \item Show reconstruction results
            \item Show pieces responsible for the matching and their locations in the reconstructed cluster
            \item Show the proposed trim frame
        \end{itemize}
    \end{itemize} 
    &
    \begin{itemize}[leftmargin=*]
        \item 2D GUI
    \end{itemize}
    \\
    \hline

    \end{tabularx}
    \label{table:JISA_puzzle}
\end{table*}

\subsection{System Overview}
This sub-section describes how the proposed JISA framework is applied to automated puzzle reconstruction, following the framework presented in Section (\ref{sec:meth}).  Table \ref{table:JISA_puzzle} lists the selections we made for puzzle solving based on the JISA framework.

\begin{figure}[t]
\centering
\includegraphics[width=3.2 in]{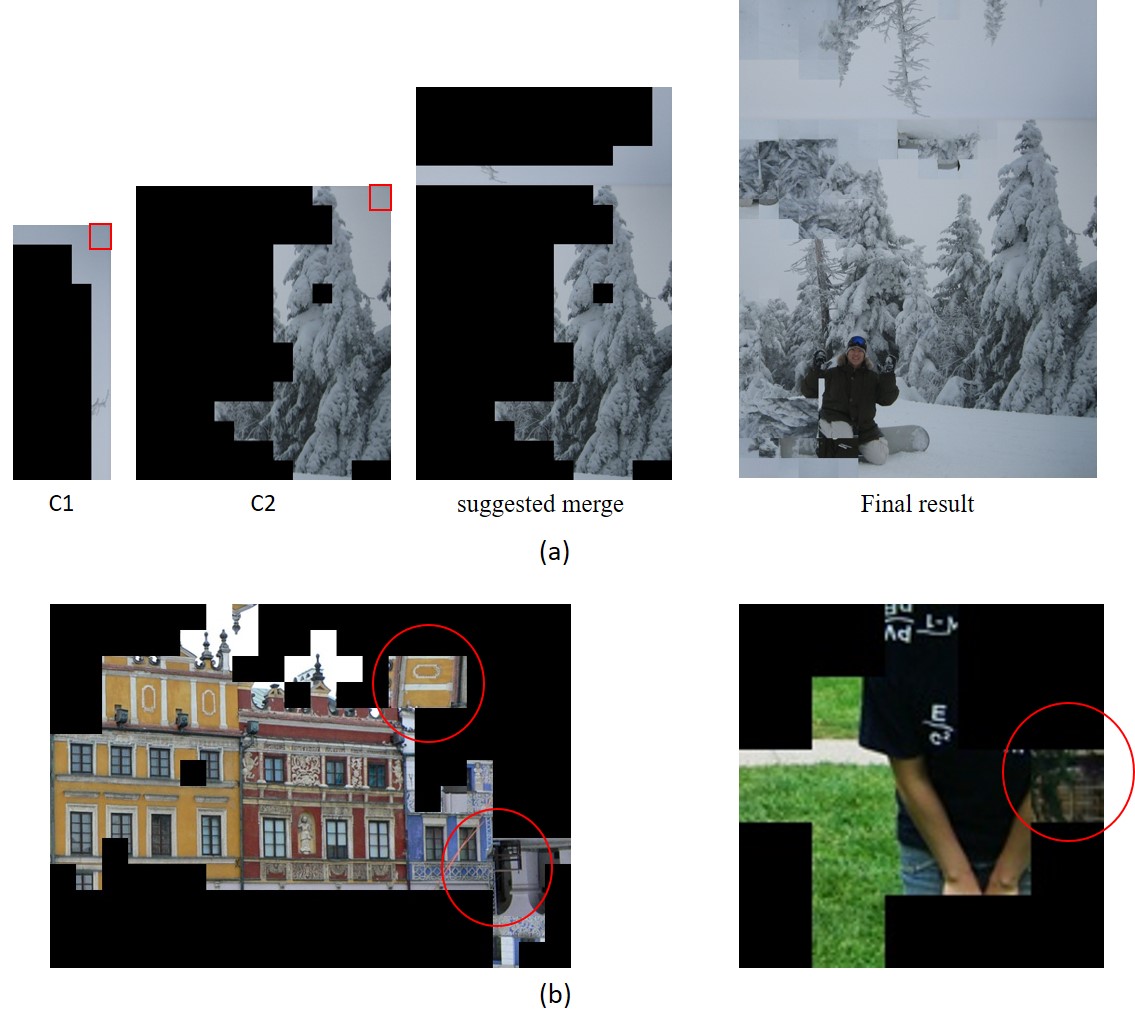}
\caption{Examples of some challenges that face autonomous greedy puzzle solvers. In (a), clusters C1 and C2 are wrongly merged based on low pairwise compatibility score; this yields a final result with low accuracy. (b) shows two examples of pieces with high textures that are misplaced by the autonomous agent.}
\label{fig:misplaced}
\end{figure}

\noindent \textbf{1. Tasks to be supervised by the human:} Based on the surveyed literature and numerical experimentation with the methods proposed in \cite{c_p5} and \cite{c_p13}, three main challenges that face greedy autonomous puzzle solvers are identified:

\begin{itemize}
    \item In cases where the pieces have low texture and highly similar colors at the edges, the pairwise compatibility score will be low and the system would match the two pieces, even though they are not adjacent in the original image. This case is referred to as a "false-positive" match, which leads to local minima and negatively affects the puzzle reconstruction accuracy. Fig. \ref{fig:misplaced}a shows an example of such cases; clusters C1 and C2 are merged together based on a suggested match of the pieces with red borders. This false-positive merge accumulated and decreased the accuracy of the final result.

    \item Although compatibility scores and priority selection were proven to be reliable \cite{c_p5}, they do guarantee correct matching between pieces even if they are rich in textures \cite{c_p33} (see examples shown in Fig. \ref{fig:misplaced}b). Moreover, if two pieces are matched based on low pairwise compatibility, but are later shown to be globally misplaced, greedy methods do not have the ability to move or rotate the misplaced piece, thus the system gets stuck in local minima and the error accumulates and diminishes the accuracy of reconstruction.
    \item The hereby adopted method reconstructs the pieces without dimension/orientation constraints. After all of the pieces are reconstructed in one cluster, the system performs a trim and fill as discussed later. This trimming may have wrong location/orientation based on the accuracy of the reconstructed image, thus the final result after re-filling the puzzle might still not match the original image.
\end{itemize} 

 Based on the above challenges, a human supervisor should be able to (1) evaluate the matching option when two pieces with low image features (not enough texture or color variation in the piece) are to be matched, (2) delete pieces that s/he finds misplaced, (3) approve the trim frame or re-locate/re-orient it after reconstruction is completed.

\noindent \textbf{2. Robot SC attributes:} From experimentation, it was realized that as the texture complexity in a puzzle piece decreases, the possibility of having a false-positive match increases. This texture complexity could be measured through the entropy of an image. In short, when the complexity of the texture in the image increases, the entropy value increases, and vice versa. Thus, the entropy of a puzzle piece is used as SC-metric. In our proposed framework implementation, when this SC-metric drops below a certain threshold that is defined experimentally, the agent asks the human supervisor to approve or decline the match decision. In case no response is received within a pre-defined time limit, the agent proceeds with the match. This no-response strategy is selected because even if merging the clusters resulted in misplaced pieces, the human can detect the error and fix it through SA-based intervention at a later stage.

\noindent \textbf{3. Human SA attributes:} since SC attributes could not be defined to capture globally misplaced pieces, the human supervisor should leverage his/her SA and judgment abilities to assess the global consistency in the reconstructed puzzle. The \textit{SA tools} proposed to enhance human SA here are summarized in showing the user the reconstruction results, highlighting the pieces responsible for merging in addition to their location in the resulted cluster, and showing the user the proposed trim frame when reconstruction is done. These tools allow the supervisor to detect misplaced pieces and relay this information back to the agent. Moreover, the user is allowed to re-locate or re-orient the trim frame if needed.

\noindent \textbf{4. JISA-based user interface:} a user-friendly GUI, shown in Fig. \ref{fig:p_GUI}(a), is developed to apply the SA tools discussed above. The GUI displays the progress of the reconstruction progress to the user, communicates with him/her the system's requests for assistance, and furnishes certain internal states that help the user intervene in the process.

\begin{figure}[b]
\centering
\includegraphics[width=3.2 in]{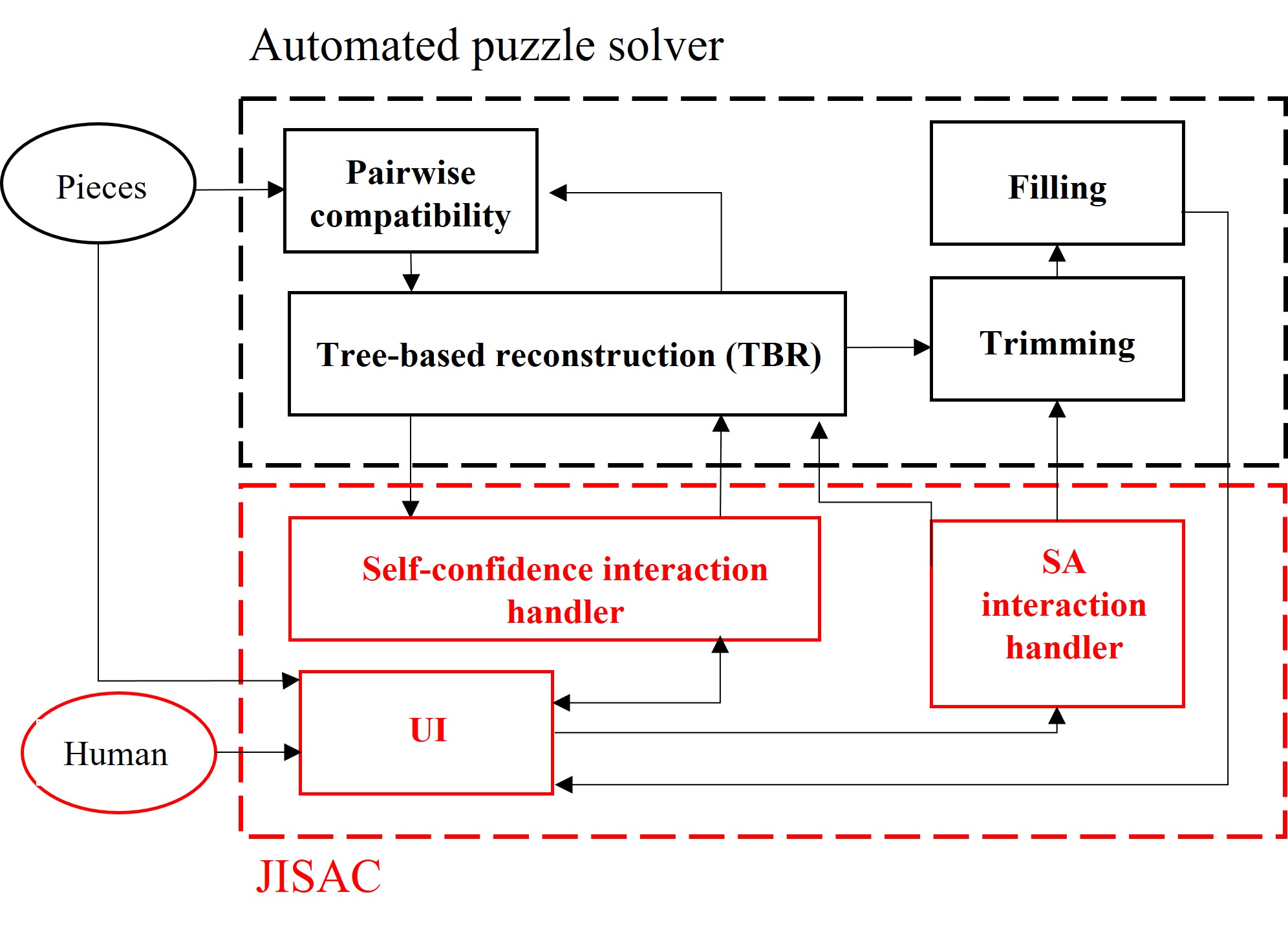}
\caption{Block diagram of the JISA-based methodology that is followed for automated jigsaw puzzle reconstruction.}
\label{fig:puzzle_meth}
\end{figure}

In this work, we are applying JISA to the automated puzzle solver presented in  \cite{c_p5} and \cite{c_p13}. This allows a human to supervise and intervene in the reconstruction of an image from a set of non-overlapping pieces. All pieces are square in shape and are placed in random locations and orientations, and only the dimensions of the final image are known to the system. Fig. \ref{fig:puzzle_meth} presents the block diagram of the JISA puzzle solver. The modules in black are adopted from  \cite{c_p5} and \cite{c_p13}, while the added JISA modules are shown in red. 
\subsubsection{\textbf{Pairwise Compatibility}}
This module calculates the pairwise compatibility function for each possible pair of pieces in the puzzle set. The MGC metric \cite{c_p13} is adopted to determine the similarity of the gradient distributions on the common boundary of the pair of pieces to be matched. Assuming that the compatibility measure $D_{LR}(x_i\ ,\ x_j)$ of two puzzle pieces $x_i$ and $x_j$, where $x_j$ is on the right side of $x_i$, is to be calculated, the color gradients in each color channel (red, green, blue) are calculated near the right edge of $x_i$ as follows:
\begin{equation}
    G_{iL} = x_i(p,P,c) - x_i(p,P-1,c),
\end{equation}
where $G_{iL}$ is the gradients array with 3 columns, $c$ represents the three color channels, and $P$ is the number of rows since the dimension of the puzzle piece (in pixels) is $P\times P$. Then, the mean distribution $\mu_{iL}(c)$ and the covariance $S_{iL}$ of $G_{iL}$ is calculated on the same side of the same piece $x_i$ as:
\begin{equation}
    \mu_{iL}(c) = \frac{1}{P} \sum_{p =1}^{P}G_{iL}(p,c).
\end{equation}
The compatibility measure $D_{LR}(x_i\ ,\ x_j)$, which calculates the gradient from the right edge of $x_i$ to the left edge of $x_j$, is calculated as:
\begin{equation}\label{eq: puzzle_DLR}
    D_{LR}(x_i\ ,\ x_j) = \sum_{p =1}^{P}(G_{ijLR}(p) - \mu_{iL})S_{iL}^{-1}(G_{ijLR}(p) - \mu_{iL})^T,
\end{equation}
where 
    $G_{ijLR} = x_j(p,1,c) - x_i(p,P,c)$.
After that, the symmetric compatibility score $C_{LR}(x_i,x_j)$ is computed as:
\begin{equation}
    C_{LR}(x_i,x_j) = D_{LR}(x_i\ ,\ x_j) + D_{RL}(x_j\ ,\ x_i),
\end{equation}
where $D_{RL}(x_j\ ,\ x_i)$ is calculated in a similar way to (\ref{eq: puzzle_DLR}). 
These steps are applied to each possible pair of pieces, for all possible orientations of each piece $({0^{\circ}}, {90^{\circ}}, {180^{\circ}}, {270^{\circ}})$.
The final step entails dividing each compatibility score $C_{LR}(x_i,x_j)$ with the second-smallest score corresponding to the same edge, resulting in the final compatibility score $C'_{LR}(x_i,x_j)$. This step ensures that significant matches of each edge has a score that is much smaller than 1, thus leading to re-prioritizing the compatibility scores in a way where pieces edges, which are more likely to be a `correct match,' are chosen by the reconstruction algorithm in the very early steps.
\subsubsection{\textbf{Tree-based Reconstruction (TBR)}}\label{sec:TBR}
\begin{figure}[t]
\centering
\includegraphics[width=3.3 in]{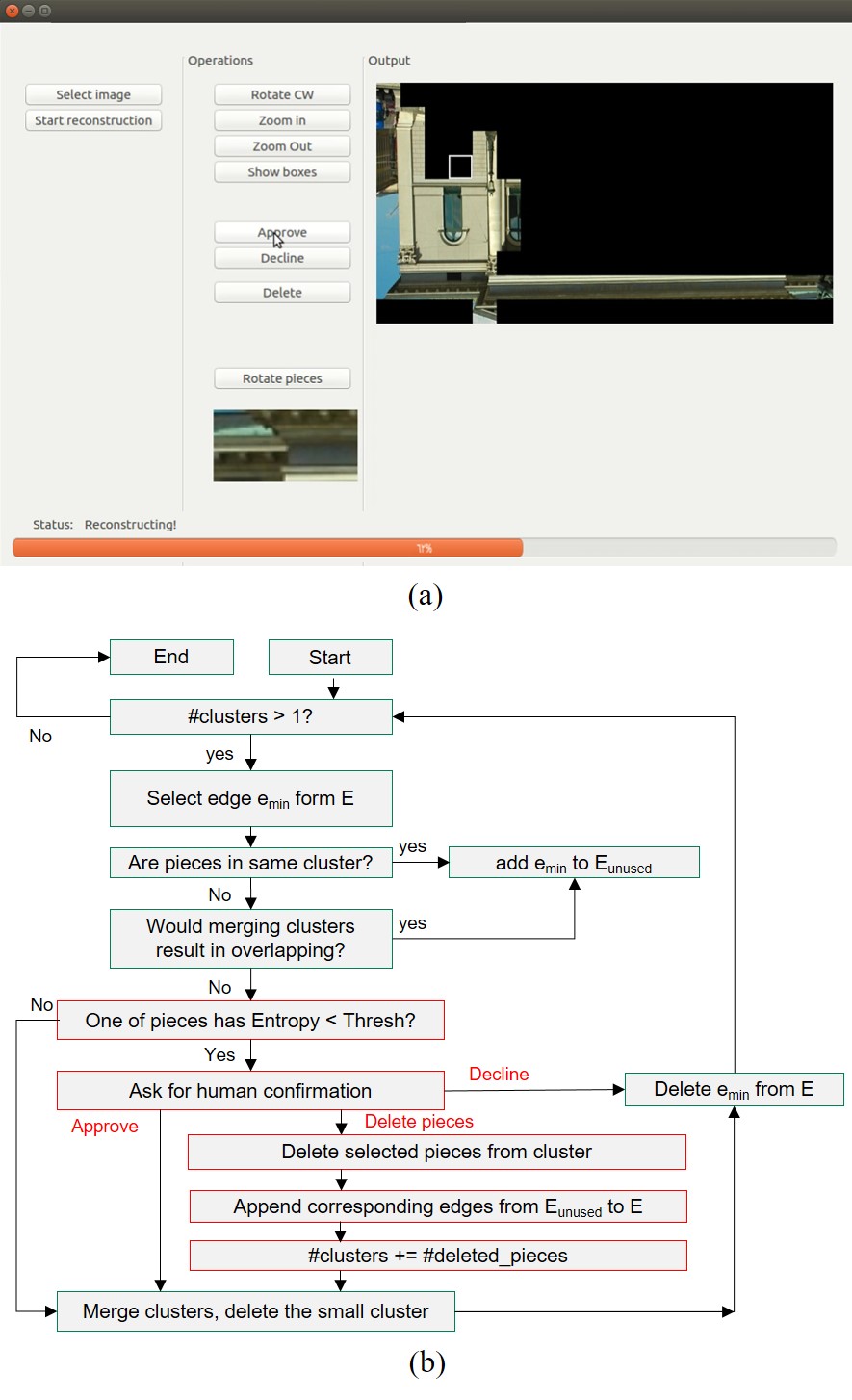}
\caption{(a) The graphical user interface (GUI) developed for applying the JISA framework to jigsaw puzzle reconstruction. (b) A flowchart showing the implementation logic of TBR, the blocks with red borders correspond to the SC and SA interaction handlers. A similar logic is used for implementing the \textit{Trimming} and \textit{Filling} modules.}
\label{fig:p_GUI}
\end{figure}

This module is responsible for reconstructing the puzzle pieces through finding a minimum spanning tree (MST) for a graph representation of the puzzle $G = (V, E)$ \cite{c_p5, c_p13}. For that, pieces are treated as vertices, and compatibility scores ($C_{LR}(x_i,x_j)$) are treated as weights of edges ($e$) in the graph. Each edge has a corresponding configuration between the two vertices (the orientation of each piece). To find the MST that represents a valid configuration of the puzzle, the method presented in \cite{c_p5, c_p13} is adopted where a modified Disjoint Set Forest (DSF) data structure is applied. The \textit{TBR} initializes with forests (clusters) equal to the number of puzzle pieces, and each forest having an individual vertex $V$ corresponding to a puzzle piece. Each forest records the local coordinates and the orientation of each member puzzle piece (vertex). A flowchart that shows the implementation logic of TBR is shown in Fig. \ref{fig:p_GUI}(b).

At the beginning of every iteration, the edge representing the lowest compatibility score ($e_{min}$) is selected in order to join the corresponding vertices in one forest. If the two vertices are already in the same forest (\textit{i.e.}, pieces belong to same cluster), or matching the pieces leads to collision in their corresponding clusters, the edge is discarded and appended to unused edges list ($E_{unused}$) . Otherwise, the two pieces, with their corresponding clusters, are sent to the \textit{SC interaction handler} module to be checked for merging. At the end of every iteration, and based on the results from the \textit{SC interaction handler} module, \textit{TBR} either discards the edge or merge the corresponding vertices into a single forest where $e_{min}$ is moved to the set of edges in this forest; here, the local coordinates and orientations of each piece are updated. Moreover, this module is responsible for applying edits requested by the human through the \textit{SA interaction handler} as discussed later. The aforementioned process repeats until all pieces are assembled in one cluster, which means that all vertices now belong to the same forest. 

\subsubsection{\textbf{Self-confidence interaction handler}}\label{sec:SC_inter}
This module represents the mechanism applied to check the self-confidence of the agent in merging two pieces with their corresponding clusters. When the two pieces (corresponding to $e_{min}$) along with their orientations are received, this module checks if either piece has an entropy below the pre-defined threshold, referred to as the confidence threshold. Our SC-metric is based on the entropy obtained from Gray Level Co-Occurrence Matrix ($GLCM$), which extracts second order statistical texture features from images. $GLCM$ is a square matrix where the number of rows and columns equals the number of gray levels in the image. Each element $GLCM(i,j)$ of the matrix is represented as a joint probability distribution function $P(i,j | d_{pixel},\theta)$, which is the probability of appearance of two pixels with values $v_i$ and $v_j$, separated by distance $d_{pixel}$ at an orientation angle $\theta$. After the GLCM is obtained, the entropy is calculated as follows:
\begin{equation}
    Entropy = - \sum_{i=o}^{M-1}{\sum_{j=o}^{M-1}{P(i,j | d_{pixel},\theta)\log_{2}P(i,j | d_{pixel},\theta)}}
\end{equation}
\noindent where $P(i,j | d_{pixel},\theta) = GLCM(i,j)$.

Since each piece of the puzzle can be placed in four different orientations, we calculate the entropy corresponding to the orientation of each piece where $\theta = 0^{\circ},  90^{\circ}, 180^{\circ}$, or $270^{\circ}$; $d_{pixel}=1$ in all calculations.

If both pieces pass this SC test, \textit{SC interaction handler} commands \textit{TBR} to proceed with merging the clusters. Otherwise, this module performs a temporary merge, shows the merging result (cluster image) to the user through the JISA-based UI, and requests human assistance in approving the merge. After requesting assistance, the \textit{SC interaction handler} awaits human response for a defined period of time. If the human approves the temporary merge, \textit{TBR} proceeds with merging. If the human declines the merge, merging is undone and the corresponding $e_{min}$ is deleted from $E$ in \textit{TBR}. In case no response is received within this time limit, \textit{TBR} proceeds with merging. 

\subsubsection{\textbf{Situation awareness interaction handler}}\label{sec:SA_inter}
In addition to approving or declining the merge upon the agent's request, the user can select pieces that s/he deems as misplaced in the resultant image based on his/her SA and judgment. When the user selects these pieces, \textit{SA interaction handler} commands \textit{TBR} to delete the corresponding vertices from the shown cluster image. Here, \textit{TBR} deletes the vertices from the displayed forest and forms new forests, using corresponding edges from $E_{unused}$, where each forest contains one vertex corresponding to one deleted piece. The vertices of the newly formed forests are connected to the other vertices in the graph through edges corresponding to the compatibility scores as discussed before. Allowing the user to inform the system about misplaced pieces is crucial since not all errors occur due to low entropy, as some pieces might have complex texture (high entropy) and still be misplaced by the system.

\subsubsection{\textbf{Trimming and Filling}}\label{sec:trim}
Trimming is applied to ensure that the final reconstructed image has the same dimensions as the original one. In this step, a frame with size equal to the original image is moved over the assembled puzzle to determine the location of the portion with the highest number of pieces; this frame is tested for both portrait and landscape orientations. When the frame with most number of pieces is defined, the \textit{Trimming} module sends the candidate frame to \textit{SA interaction handler}, which shows the user this frame superposed over the reconstructed image. The user here can approve the frame location/orientation or edit it. After the frame is set, all pieces outside this frame are trimmed from the image and sent to the \textit{Filling} module. Here, the trimmed pieces are used to fill the gaps in the reconstructed image. Filling starts from gaps with the highest number of neighbors and continues until all gaps are filled. Each gap is filled by a piece whose edges has best compatibility score with the adjacent edges corresponding to the gap's neighbor pieces.

\subsection{Implementation and Experiments}\label{sec:p_imp}
The JISA framework is implemented in Python3, and the GUI is developed in pyQT. The GUI, shown in Fig. \ref{fig:p_GUI}(a), displays the results of the reconstruction in real-time (a resultant cluster image is shown in the `output' in the GUI), the percentage of completion of the reconstruction process, in addition to the logs/errors/warnings. Moreover, and to help the user decide whether to decline the merge or delete misplaced pieces, the GUI shows the user the two pieces that are responsible of merging, and the location of these two pieces in the resultant cluster image. 

Through this GUI, the user can: choose the image to be reconstructed, start reconstruction process, and perform operations on the shown cluster image. These operations are: rotate, zoom in/out, approve/decline, and delete pieces from the cluster. When the final cluster is obtained, the GUI shows the proposed trimming frame to the user for approval or re-location.

To validate the proposed JISA framework in reconstructing jigsaw puzzles, experiments were conducted on the MIT dataset published in \cite{c_p1}. This dataset is formed of 20 images and is used as a benchmark in most of the available references in the related literature. To assess the results, two types of accuracy measures are calculated. The \textbf{direct comparison metric} compares the exact position and orientation of each piece with the original image and calculates the accuracy as a percentage. A drawback of this metric is that if the reconstructed image has its pieces slightly shifted in a certain direction, it will report a very low score, and even Zero. The \textbf{neighbor comparison metric} compares the number of times that two pieces, which are adjacent (with the same orientation) in the original image, are adjacent in the final solution. This metric shows more robustness against slightly shifted pieces. 

In each experiment, the user selects an image from the set through the GUI, then starts the reconstruction process. The human supervisor is instructed to decline the merge option if s/he is not totally confident that this match is correct. Moreover, the user is instructed to delete misplaced pieces whenever detected. Here, it is important to mention that the user is assumed to have a detailed knowledge about the GUI and how the algorithm works. Three runs were performed on each image, resulting in a total of 60 tests. Table \ref{table:puzzle_results} shows the results obtained by the JISA system, in addition to results from four different previous works. All results are based on square puzzle pieces with unknown locations and orientations. The number of pieces per puzzle is 432 and piece size in pixels is $28 \times 28$. The obtained results show that the JISA framework outperforms the other approaches in both direct and neighbor comparison metrics.

\begin{figure}[!t]
\centering
\includegraphics[width=3.2 in]{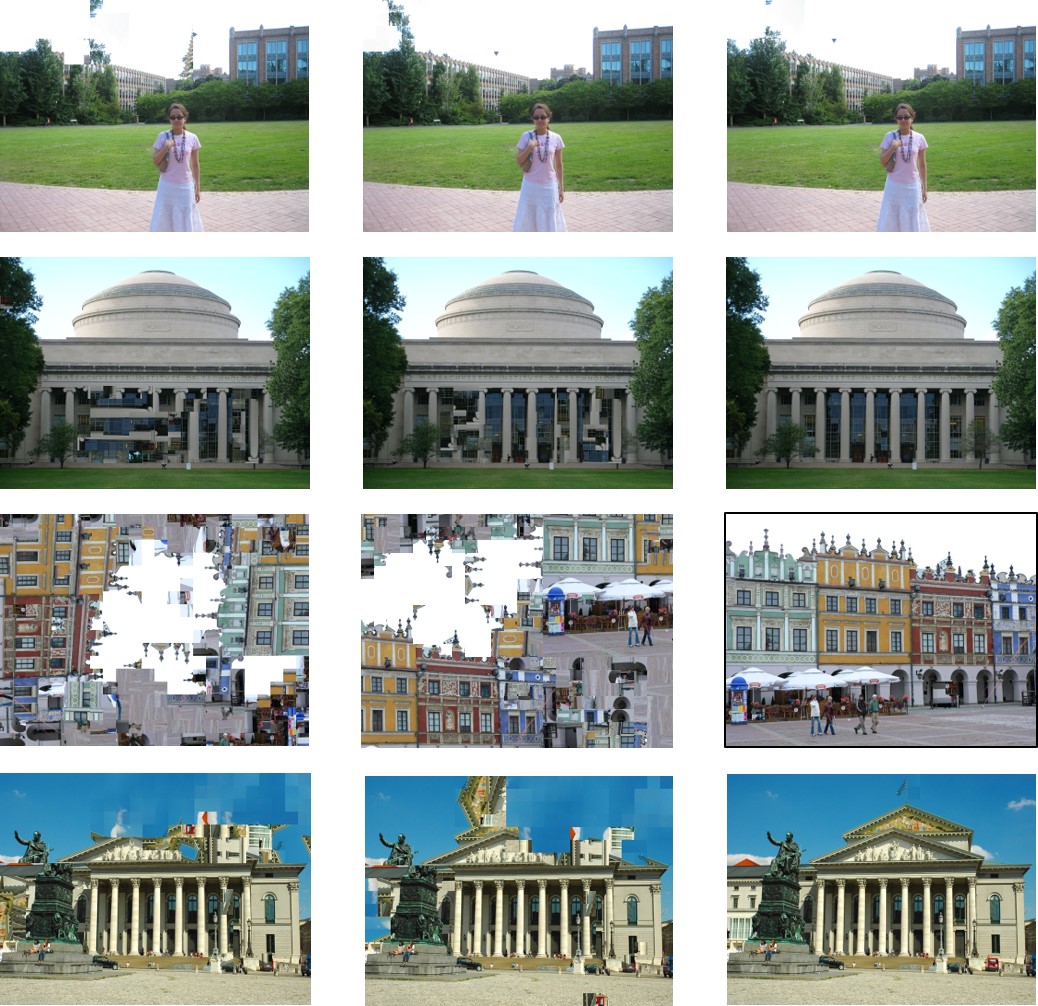}
\caption{Visual results of conducting. The left column shows results using the approach in \cite{c_p13}, the middle column shows the results obtained through running the proposed system but the user was allowed only to approve/decline merge options, and the column to the right shows the results obtained by the proposed JISA puzzle solver.}
\label{fig:comparison}
\end{figure}

To further show the effect of having JISA, four experiments are conducted in which the supervisor is only allowed to approve or decline the merge result when the system's accuracy drops. This means that the supervisor can only intervene upon the system's request; so the system cannot benefit from the supervisor's better SA and judgment to correct errors that it cannot detect. Visual results are shown in Fig. \ref{fig:comparison}. The first column (left) shows results of reconstruction from \cite{c_p13}, the second column (middle) shows results of the user \textbf{only} approving/declining upon the system's request, while the third column (right) shows results of the proposed JISA system. As expected, these results demonstrate that having a JISA framework, where the supervisor can intervene based on his/her SA \textbf{and} the system's request, results in superior performance as compared to a fully autonomous system or a system that only asks for human assistance based on its self-confidence. 
\begin{table}[!t]
\centering
\scriptsize
\caption{Comparison of reconstruction performance between the JISA framework and four state-of-the-art approaches.}
\begin{tabular}{|c|c|c|c|c|} 
\hline
 Approach & Year & Direct Metric & Neighbor Metric 
\\
\hline
 JISA framework &2021 & $96.5\%$ & $97.2\%$
 \\
 MTS with DSF \cite{c_p13}  & 2016 & $88.5\%$ & $92.9\%$
 \\
 Linear Programming \cite{c_p25} & 2015 & $95.3\%$ & $95.6\%$
 \\
 Loop Constraints \cite{c_p41} & 2014 & $94.7\%$ & $94.9\%$
 \\
 Constrained MST \cite{c_p5}  & 2012 & $82.2\%$ & $90.4\%$
 \\
 \hline
 \end{tabular}
\label{table:puzzle_results}
\end{table}
\section{Discussion and Conclusion}
The aim of this paper is to propose and validate a joint-initiative supervised autonomy framework for human-robot interaction.  Unlike other frameworks that focus on the allocation of functions between humans and robots, JISA is proposed to extend the spectrum of HRI to autonomous systems where it is not feasible for a human to `completely takeover' the automated mission tasks. In the proposed JISA framework, the autonomous agent (robot) performs tasks in an autonomous fashion, but can ask a human supervisor for assistance based on its self-confidence. In addition, the human can intervene and influence the autonomous sensing, planning, and acting based on his/her SA. This framework aims to overcome autonomy challenges and enhance its performance by combining the cognition, flexibility, and problem solving skills of humans with the the strength, endurance, productivity, and precision of robots. As a proof of concept, the proposed JISA framework is applied in two different systems: collaborative grid-based SLAM, and automated jigsaw puzzle re-construction. In both systems, the following are defined: (1) the challenges and limitations affecting full autonomy performance, (2) the confidence measure that triggers the robot’s request for human assistance,  (3) the type and level of intervention that the human can perform.  In addition, an augmented reality interface (for SLAM) and 2D graphical user interface (for puzzle reconstruction)  are custom-designed to enhance the human situation awareness and communicate information and requests efficiently. 

Through experimental validation, it was shown that applying the JISA framework in fully autonomous systems can help overcome several autonomy limitations and enhance the overall system performance. In fact, JISA outperformed full autonomy in both implemented systems. In collaborative SLAM, post processing of grid maps was eliminated and the JISA system produced more accurate maps in less number of trials and less run-time for each trial. In automated puzzle reconstruction, results showed that the JISA system outperforms both fully autonomous systems and systems where the human merely intervenes (accept/reject) upon the agent’s requests only. 

Since this is a first step towards a joint-initiative supervised autonomy, we are aware of some limitations that need to be addressed in any future evolution of JISA. First, the cost of each interaction over its benefit is not evaluated in this work. This is a crucial component that could be included in the SC/SA attributes to reason about when it is best to request/get assistance through measuring: (1) the cost of interrupting the human supervisor, which is most important when the human is supervising multiple robots; and (2) the cost of waiting for human response, which is important in missions where the robot needs to make rapid decisions and actions. Another limitation is that JISA assumes the human to be available and ready to provide assistance whenever needed, which is not always the case. This assumption could be relaxed by including a module to estimate the human availability and evaluate his/her capability to provide the needed help. Third, the number of help requests within a mission could be reduced if the robot has the learning-from-interaction capability. Agent learning would lead to extending the types of human assistance to include providing demonstrations, information, and preemptive advice. Finally, to relax the assumption that the human always has superiority over the robot decisions, a human performance evaluation module could be included to reason about whether to accept the human assistance, negotiate it, or refuse it.

In future work, we will be studying more SC metrics/events (other than $N_{eff}$ and Entropy) in both POC applications presented. This will help in evaluating the effectiveness and optimality of the proposed attributes. In addition, we aim to perform more tests on both applications through a group of human subjects. The goal of these tests is to avoid any biased results and study certain factors that might influence the performance of human supervisors such as awareness, workload, skills, trust, and experience. Moreover, and to better validate the general applicability and efficiency of the proposed JISA framework, we aim to apply it in a new task of physical robot 3D assembly.

\section{Acknowledgments}
This work was supported by the University Research Board (URB) at the American University of Beirut.
\bibliography{references}
\end{document}